\documentclass{article} 
\usepackage{iclr2025_conference,times}
\iclrfinalcopy 

\usepackage{amsmath,amsfonts,bm}

\def\eqref#1{equation~\ref{#1}}

\def\1{\bm{1}}

\DeclareMathAlphabet{\mathsfit}{\encodingdefault}{\sfdefault}{m}{sl}
\SetMathAlphabet{\mathsfit}{bold}{\encodingdefault}{\sfdefault}{bx}{n}

\usepackage{hyperref}
\usepackage{url}
\usepackage{bbding}
\usepackage{graphicx}
\usepackage{amsmath}
\usepackage{amssymb}
\usepackage{booktabs}
\usepackage{bbm}
\usepackage{colortbl}
\usepackage{flushend}
\usepackage{bm}
\usepackage{multirow}
\usepackage{makecell}
\usepackage{tikz}
\usepackage{comment}
\usepackage{amsmath,amssymb} 
\usepackage{color}
\usepackage{colortbl}
\usepackage{tikz}
\usepackage{caption}
\usepackage{hyperref}
\usepackage{url}
\usepackage{times}
\usepackage{epsfig}
\usepackage{graphicx}
\usepackage{amsmath}
\usepackage{amssymb}
\usepackage{bbm}
\usepackage{multirow}
\usepackage{wrapfig}
\usepackage[capitalize]{cleveref}
\crefname{section}{Sec.}{Secs.}
\Crefname{section}{Section}{Sections}
\Crefname{table}{Table}{Tables}
\crefname{table}{Tab}{Tabs.}
\usepackage{subcaption}

\newlength\savewidth

\ExplSyntaxOn
\newcommand\latinabbrev[1]{
	\peek_meaning:NTF . {
		#1\@}%
	{ \peek_catcode:NTF a {
			#1.\@ }%
		{#1.\@}}} 
\ExplSyntaxOff

\def\eg{\latinabbrev{e.g}}

\def\ie{\latinabbrev{i.e}}

\usepackage[textsize=tiny]{todonotes}

\definecolor{verylightgray}{RGB}{234, 250, 254}
\definecolor{fgreen}{RGB}{15, 159, 94}

\title{ControlVAR: Exploring Controllable Visual Autoregressive Modeling}

\author{%
  Xiang Li$^1$\thanks{xl6@andrew.cmu.edu}, Kai Qiu$^1$, Hao Chen$^1$, Jason Kuen$^2$, Zhe Lin$^2$, Rita Singh$^1$, Bhiksha Raj$^{1,3}$ \\
  $^1$Carnegie Mellon University, $^2$Adobe Research, $^3$MBZUAI\\
}

\begin{document}

\maketitle

\begin{figure}[h]
    \centering
    \includegraphics[width=\linewidth]{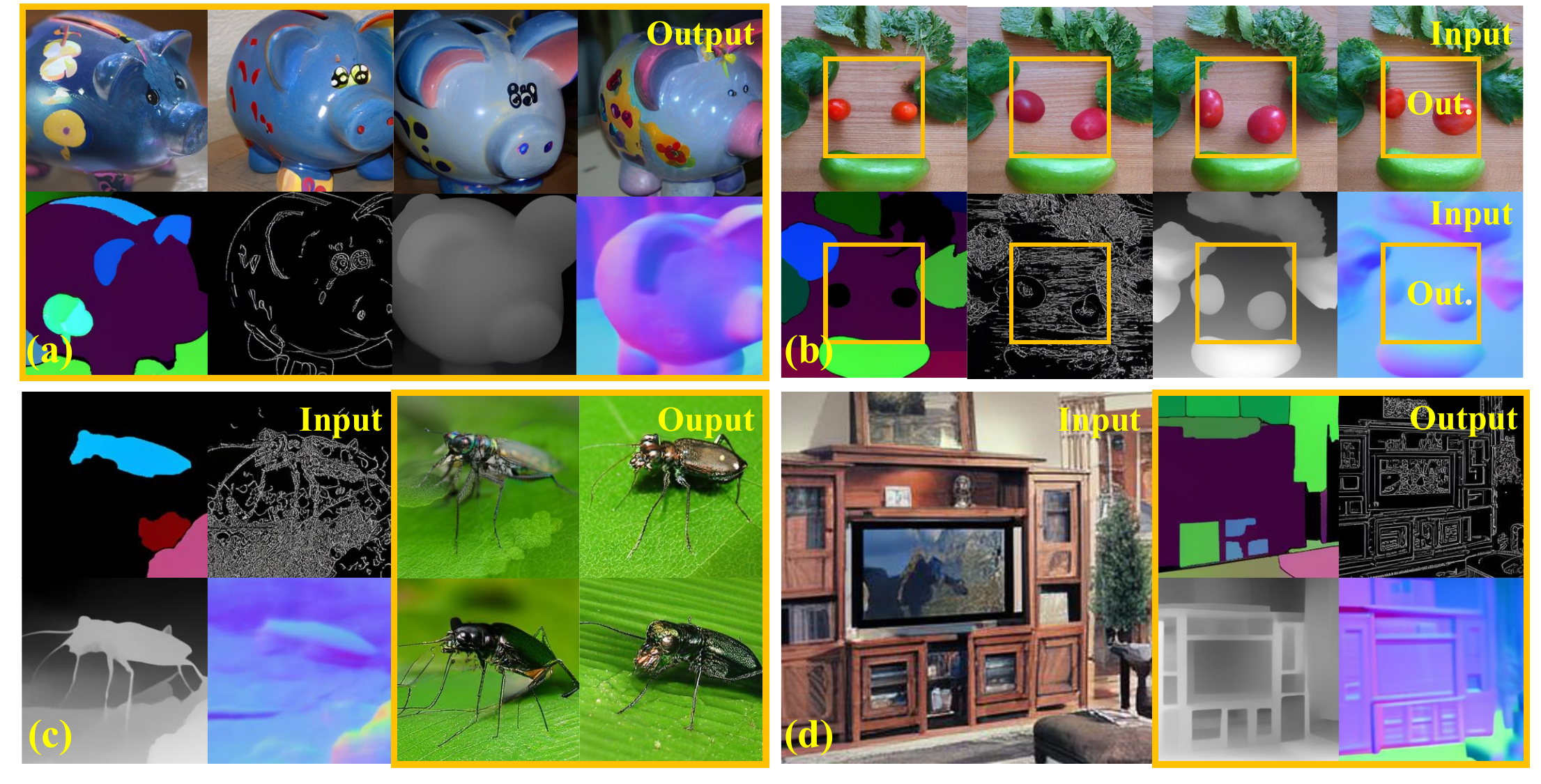}
    \caption{Visualization of ControlVAR for (a) joint control-image generation, (b) 
    joint control-image completion, (c) control-to-image generation, and (d) image-to-control prediction (visual perception tasks). The {\color{yellow}{yellow}} boxes denote the predicted images/controls.}
    \label{fig:teaser0}
\end{figure} 

\begin{abstract}
Conditional visual generation has witnessed remarkable progress with the advent of diffusion models (DMs), especially in tasks like control-to-image generation. However, challenges such as expensive computational cost, high inference latency, and difficulties of integration with large language models (LLMs) have necessitated exploring alternatives to DMs. This paper introduces ControlVAR, a novel framework that explores pixel-level controls in visual autoregressive (VAR) modeling for flexible and efficient conditional generation. In contrast to traditional conditional models that learn the conditional distribution, ControlVAR jointly models the distribution of image and pixel-level conditions during training and imposes conditional controls during testing. 
To enhance the joint modeling, we adopt the next-scale AR prediction paradigm and unify control and image representations. 
A teacher-forcing guidance strategy is proposed to further facilitate controllable generation with joint modeling. Extensive experiments demonstrate the superior efficacy and flexibility of ControlVAR across various conditional generation tasks against popular conditional DMs, \eg, ControlNet and T2I-Adaptor. Code: \url{https://github.com/lxa9867/ControlVAR}.
\end{abstract}
\section{Introduction}
In recent years, conditional image generation~\cite{zhang2023adding,mou2023t2i,esser2021taming,tian2024visual,nam2024dreammatcher} has attracted great attention and there have been significant advancements in text-to-image generation~\cite{rombach2021highresolution,chang2023muse,gal2022image}, image-to-image generation~\cite{zhang2023adding,mou2023t2i,ruiz2023dreambooth}, and even more complex tasks~\cite{nam2024dreammatcher,li2023completing,li2024paintseg}. Most recent approaches, \eg, ControlNet \cite{zhang2023adding}, leverage the powerful diffusion models (DMs) \cite{rombach2021highresolution,peebles2023scalable} to model the large-scale image distribution and incorporate additional controls with classifier-free guidance~\cite{ho2022classifier}. However, the inherent nature of the diffusion process imposes many challenges for the diffusion-based visual generation: (1) the computational cost and inference time are significant due to the iterative diffusion steps \cite{song2020denoising,ho2020denoising} and (2) the incorporation in mainstream intelligent systems, \ie, large language models (LLMs) \cite{touvron2023llama,achiam2023gpt}, is intricate due to the representation difference. This motivates the community to find a replacement for DMs for high-quality and efficient visual generation in the era of LLMs. 

Inspired by the success of autoregressive (AR) language modeling \cite{touvron2023llama,achiam2023gpt}, AR visual modeling \cite{esser2021taming,tian2024visual} has been investigated as a counterpart to DMs given its strong scalability and generalizability \cite{tian2024visual,bai2023sequential}. Several inspiring works, \eg, VQGAN \cite{esser2021taming}, DALL-E \cite{ramesh2021dalle} and VAR \cite{tian2024visual}, have demonstrated promising image generation results with AR modeling. Nevertheless, compared to the prosperity of conditional DMs \cite{zhang2023adding,mou2023t2i,chen2022re,xu2023versatile,qin2023unicontrol,ju2023humansd}, visual generation with conditional AR modeling \cite{zhan2022auto,esser2021taming} remains significantly under-explored. Different from DMs, where all the pixels are modeled simultaneously, AR models are characterized by modeling sequential values based on their corresponding previous ones. This AR approach naturally leads to a conditional model, providing potential flexibility when incorporating additional controls. To leverage this property, teacher forcing is a popular approach that controls AR prediction by replacing partially predicted tokens with ground truth ones \cite{esser2021taming}. 
Thanks to this nature of AR modeling, we found that highly flexible conditional generation can be achieved by teacher forcing partial AR sequence with proper model designs.


\begin{figure}[t]
    \centering
    \includegraphics[width=\linewidth]{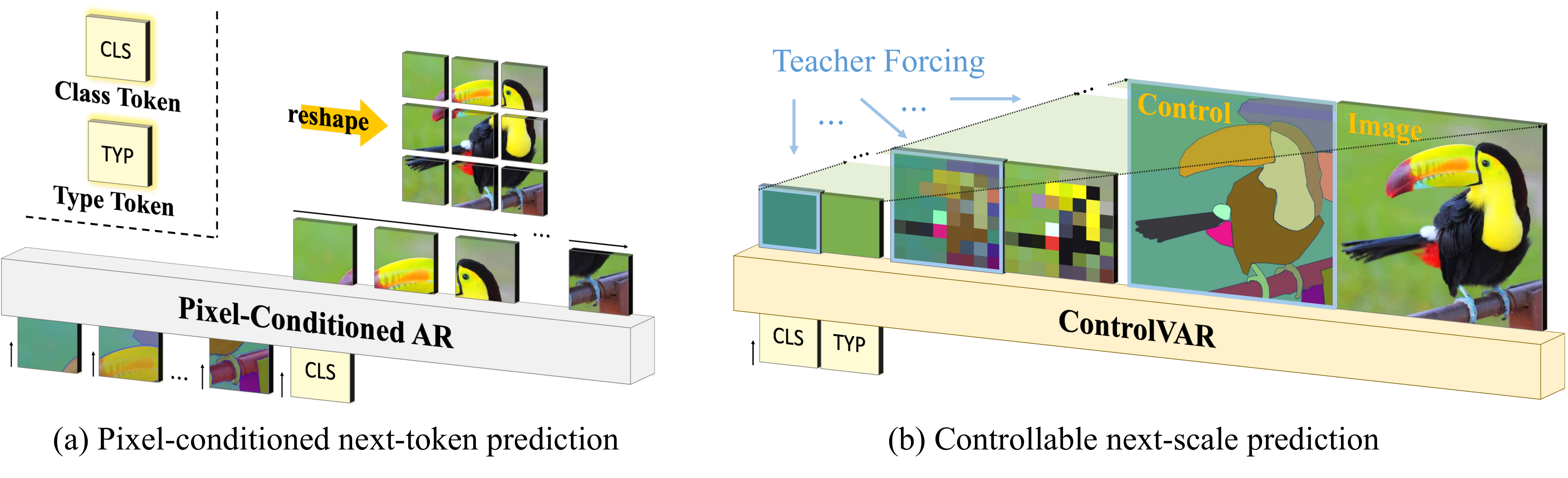}
    \vspace{-0.6cm}
    \caption{In contrast to previous methods \cite{esser2021taming,zhan2022auto} that leverage prefix conditional tokens to impose controls, ControlVAR jointly models the pixel-level controls and image during training and conducts the conditional generation tasks during testing with the teacher forcing. Class and type tokens provide semantic and control type (mask, canny, depth and normal) information respectively.}
    \label{fig:teaser}
\end{figure}

In this paper, we explore the \textbf{Control}lable \textbf{V}isual \textbf{A}utoreg\textbf{R}essive modeling with both token-level and pixel-level conditions. A new conditional AR paradigm, ControlVAR is introduced, which permits a highly flexible conditional image generation by embracing the next-scale prediction of joint control and image (\cref{fig:teaser}(b)). Previous wisdom \cite{zhan2022auto,esser2021taming} typically utilizes prefix conditions (\cref{fig:teaser}(a)) and mainly model images from raw pixel space in an AR manner. Differently, we notice that if we jointly model the control and image, the learned joint prediction can be easily guided by teacher forcing during inference. On the one hand, we unify the control and image representations and reformulate the sequential variables for the AR process to enable effective joint modeling. On the other hand, by analyzing the modeled probabilities, we introduce an effective sampling strategy, named teacher forcing guidance (TFG) to facilitate conditional sampling. Remarkably, a single ControlVAR model trained via TFG is capable of multiple meaningful tasks with different input-output combinations between control and image:
(a) joint control-image generation
, (b) control/image completion
, (c) control-to-image generation
, (d) image-to-control generation
, as demonstrated in \cref{fig:teaser0}. Beyond the image-control tasks that are jointly modeled during training, we observe that ControlVAR also emerges capabilities for unseen tasks, \eg, control-to-control generation, further enhancing its flexibility and versatility.
Our contribution can be summarized in three-fold:

\begin{itemize}
    \item We present ControlVAR, a novel framework for controllable autoregressive image generation with strong flexibility for heterogeneous conditional generation tasks.
    \item We unify the image and control representations and reformulate the conditional generation process to jointly model the image and control during training. To perform conditional generation during 
    inference, we introduce teacher-forcing guidance (TFG) that enables controllable sampling.
    \item We conduct comprehensive experiments to investigate the impacts of each component of ControlVAR and demonstrate that ControlVAR outperforms powerful DMs methods, \eg, ControlNet and T2I-Adapter on controlled image generation across several pixel-level controls, \ie, mask, canny, depth and normal.
\end{itemize}

\section{Related Works}
\subsection{Diffusion-based Image Generation}
\label{sec:related-diffusion}
The evolution of diffusion models, initially introduced by Sohl-Dickstein et al. \cite{sohl2015deep} and later expanded into image generation using fixed Gaussian noise diffusion processes \cite{ho2020denoising,song2020score}, has witnessed significant advancements driven by various research efforts. Nichol et al. \cite{nichol2021improved} and Dhariwal et al. \cite{dhariwal2021diffusion} proposed techniques to enhance the effectiveness and efficiency of diffusion models, paving the way for improved image generation capabilities. Notably, the paradigm shift towards modeling the diffusion process in the latent space of pre-trained image encoders as a strong prior \cite{van2017neural,esser2021taming} rather than raw pixels spaces \cite{vahdat2021score,rombach2022high,peebles2023scalable} has been instrumental in achieving high-quality image generation with reasonable inference speed. This approach has led to the development of foundational diffusion models such as Glide \cite{nichol2021glide}, Cogview \cite{ding2021cogview,ding2022cogview2,zheng2024cogview3}, Make-a-scene \cite{gafni2022make}, Imagen \cite{saharia2022photorealistic}, DALL.E \cite{ramesh2021zero}, Stable Diffusion \cite{stablediffusion}, MidJourney \cite{midjourney}, SORA \cite{sora}, among others, which are often pre-trained on large-scale data with conditions, typically text \cite{gordon2023mismatch,webster2023duplication,elazar2023s,chen2024catastrophic}. Recent advancements include consistency models derived from diffusion models \cite{song2023consistency,song2023improved,luo2023latent}, enabling generation with reduced inference steps. These foundational diffusion models have not only opened doors to novel downstream applications like Text inversion \cite{gal2022image}, DreamBooth \cite{ruiz2023dreambooth}, T2I-Adapter \cite{mou2023t2i}, ControlNet \cite{zhang2023adding}, but also inspired a plethora of research in controllable generation \cite{meng2021sdedit,brooks2023instructpix2pix,huang2023region,tumanyan2023plug,voynov2023sketch,huang2024diffstyler,huang2023composer,bashkirova2023masksketch,bar2023multidiffusion,li2023gligen,qi2023unigs,zhan2022auto} and other innovative areas. 

\subsection{Autoregressive Image Generation.}
Unlike diffusion-based models that typically leverage continuous image representation, autoregressive models \cite{huang2023not,esser2021taming,van2016conditional,tian2024visual} leverage discrete image tokens. An image tokenizer \cite{esser2021taming,yu2023language,yu2024spae,huang2023towards,ge2023planting} is utilized to encode the image into a sequence of discrete tokens. VQGAN \cite{esser2021taming} first patches the image and then employs a vector-quantization approach to discretize the image features. Following this paradigm, a series of following-up works improve the image tokenization by using more powerful quantization operations \cite{huang2023not,lee2022autoregressive,yu2023language}, reformulating the image representation \cite{tian2024visual,tschannen2023givt} and modifying the network architecture \cite{yu2021vector,razavi2019generating}. With the discrete tokens, a transformer structure \cite{radford2019language} is leveraged to model the image token sequences. RQ-GAN \cite{lee2022autoregressive} improves the modeling by incorporating a hierarchy design and MQ-VAE \cite{huang2023not} further utilizes StackTransformer to enhance the spatial focus. MUSE \cite{chang2023muse} is a large-scale pre-trained text-to-image model where a low-resolution image is first generated followed by a super-resolution transformer to refine the image. Recently, VAR \cite{tian2024visual} introduced a new next-scale autoregressive prediction paradigm where the image representation is shifted from patch to scales. The new representation is featured with the maintenance of spatial locality and much lower computational cost. In this paper, we follow the next-scale autoregressive paradigm and explore the incorporation of additional controls into the modeling process.

\subsection{Conditional Image Generation}
Though significant progress has been made in generating highly realistic images from textual descriptions, describing every intricate detail of an image solely through text poses challenges. To overcome this limitation, researchers have explored alternative approaches using various additional inputs to effectively control image and video diffusion models. These inputs encompass bounding boxes \cite{li2023gligen,yang2022reco}, reference object images \cite{ruiz2023dreambooth,Li2023BLIP-Diffusion}, segmentation maps \cite{Gafni2022Make-A-Scene,Avrahami2023SpaText,zhang2023adding}, sketches \cite{zhang2023adding}, and combinations thereof \cite{kim2023diffblender,qin2023unicontrol,zhao2024uni,wang2024videocomposer,mizrahi20244m,nam2024dreammatcher,zhou2023customization}. However, fine-tuning the vast array of parameters in these diffusion models can be computationally intensive. To address this, methods like ControlNet \cite{zhang2023adding} have emerged, enabling conditional control through parameter-efficient training strategies \cite{zhang2023adding,simoryu_lora_diffusion,mou2023t2i}. Notably, X-Adapter \cite{ran2023x} innovatively learns an adapter module to adapt ControlNets pre-trained on smaller image diffusion models (e.g., SDv1.5) for larger models (e.g., SDXL). SparseCtrl \cite{guo2023sparsectrl} takes a different approach, guiding video diffusion models with sparse conditional inputs, such as few frames instead of full frames, to mitigate the data collection costs associated with video conditions. However, the implementation of SparseCtrl necessitates training a new variant of ControlNet from scratch, as it involves augmenting ControlNet with an additional channel for frame masks.
Beyond traditional conditional image generation, the in-context learning capability of conditional models has also been explored \cite{safaee2023clic,mizrahi20244m,bai2023sequential,zhang2024makes}. LVM \cite{bai2023sequential} investigates the scaling learning capability of a large vision model without any linguistic data. 4M \cite{mizrahi20244m} investigate the large-scale visual generation with multimodal data using masked image modeling. Different from previous works which are mainly focusing on diffusion models, we aim to explore adding additional control to the autoregressive visual generation process.

\section{ControlVAR}
ControlVAR is an autoregressive Transformer \cite{vaswani2017attention} framework for conditional image generation tasks, using the following as conditions: 
image $I\in\mathbb{R}^{3\times H\times W}$, pixel-level control $C\in\mathbb{R}^{3\times H\times W}$ and token-level control $c\in\mathbb{R}^{D}$ where $H,W$ and $D$ denotes the image size and dimension of control token respectively. We denote the set of $N$ different types of controls as $\mathcal{C}=\{C_n\}_{n \in [N]}$.

\paragraph{Problem formulation.}

Prior conditional approaches \cite{zhang2023adding,tian2024visual} have often utilized distinct models for individual control type $C$, learning a conditional distribution in the form of $p(I|C,c)$, where each image $I$ is encoded as a sequence of discrete tokens of length $T$, denoted as $(x_1,x_2,\dots,x_T)$. By employing autoregressive (AR) modeling, we can rewrite the conditional probability $p(I|C,c)$ as
\begin{equation}
p(I|C,c)=p(x_1,x_2,\dots,x_T|C,c)=\prod_{t=1}^Tp(x_t|x_{<t},C,c)
\end{equation}
where each image token $x_t$ is conditioned on previous ones $x_{<t}$ at position $t$ and prefix controls $C,c$. 

In this paper, we consider $N$ different controls and reformulate the conditional AR generation to model the joint distribution $p(I,\mathcal{C}|c)$ during training. Specifically, we uniformly sample one control $C\in\mathcal{C}$ at each training iteration and leverage an additional type token $c_t$ to convey the control type information. Assuming the control tokens are of the same length as the image (which we will show in the next section), we represent it as a sequence of discrete tokens $C=(y_1,y_2,\dots,y_T)$. To jointly model the image and control while not losing the autoregressive properties, we group the image and control tokens as $r_t=(x_t,y_t)$ and model the joint distribution as:
\begin{equation}
p(I,C|c,c_t)=p\left((x_1,y_1),(x_2,y_2),\dots,(x_T,y_T)|c,c_t\right)=\prod_{t=1}^Tp(r_t|r_{<t},c,c_t).
\end{equation}
For inference, we introduce an innovative approach inspired by teacher forcing, which replaces the predicted token with the ground truth to perform conditional generation tasks. We will discuss the representation of $r_t$ in \cref{sec:representation}, joint control-image AR modeling in \cref{sec:joint}, and conditional generation during inference in \cref{sec:tfg}.

\subsection{Unified Image and Control Representation.}
\label{sec:representation}

Images are generally represented in RGB, which is different from how pixel-level controls (\eg, mask, canny, and depth) are represented. Although using the original representation of respective controls may be beneficial for information preservation, doing so would lead to a larger vocabulary size of the predicted tokens thus hindering effective AR modeling. To this end, we aim to represent the controls with the same RGB representation of images. 

\paragraph{Control representation.}

\begin{wrapfigure}{r}{0.5\linewidth}
  \centering
  \vspace{-0.2cm}
  \includegraphics[width=\linewidth]{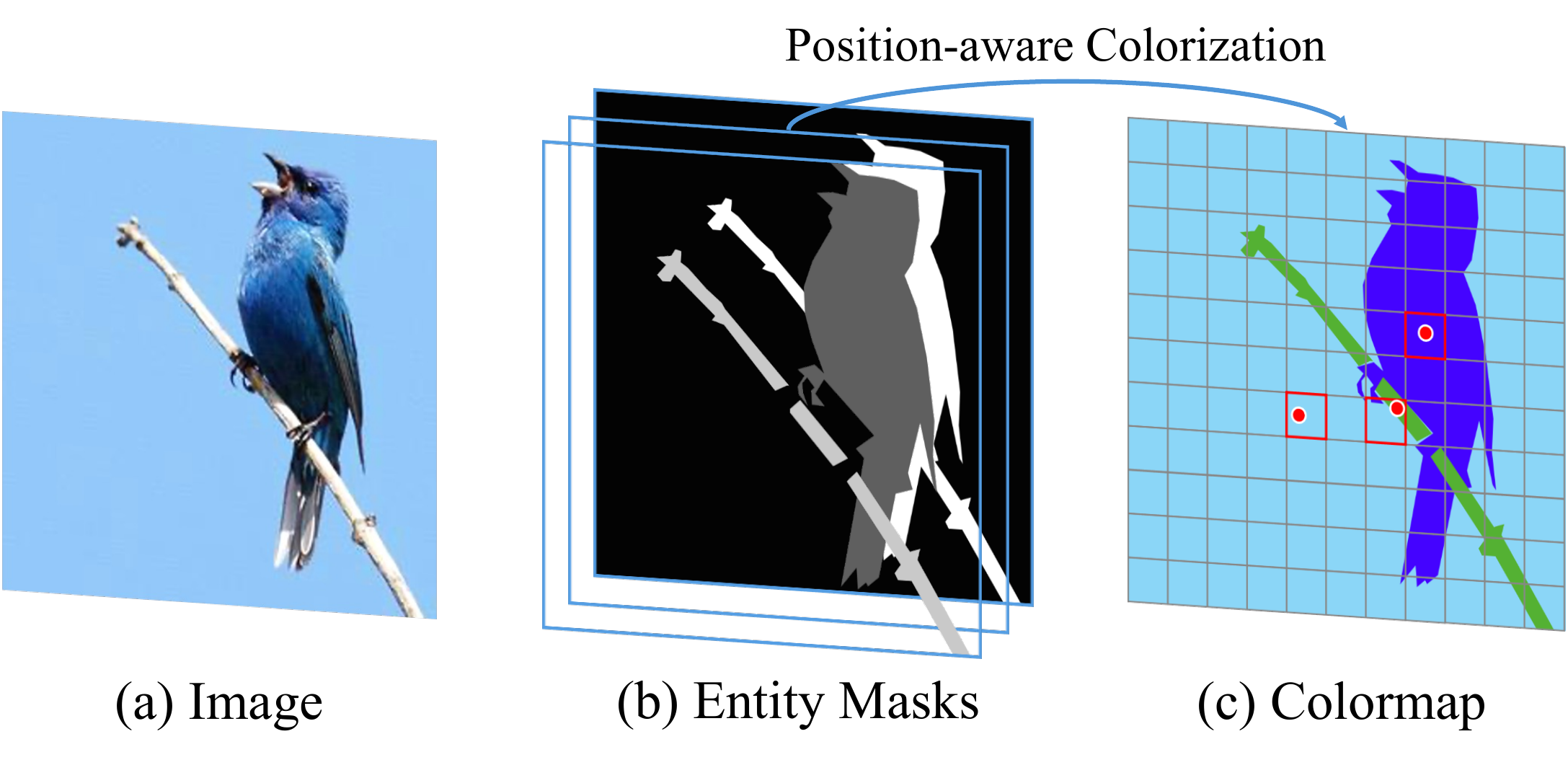} 
  \caption{Illustration of colormap representation.}
  \label{fig:colormap}
\end{wrapfigure}

We consider four popular control types - entity mask, canny, depth, and normal in this paper. We notice that canny, depth, and normal can be easily converted to RGB by using simple transformations \cite{zhang2023adding}. 
However, entity segmentation masks
$M\in\{0,1\}^{N\times H\times W}$ which comprises $N$ class-agnostic binary masks (\cref{fig:colormap}(b)) cannot be easily converted. 
Inspired by SOLO \cite{wang2020solo}, we leverage a position-aware color map to encode the binary masks $M$ into a colormap $M^\prime\in[0,255]^{3\times H\times W}$. 
To better distinguish the color difference, we select 5 candidate values $\{0, 64, 128, 192, 255\}$ from each RGB channel and combine them to $124=5^3-1$ colors ($(0,0,0)$ is preserved for background). 
To apply the colormap, as shown in \cref{fig:colormap}, we divide the image into $n_h\times n_w$ regions where each region represents a corresponding color. We calculate the centeredness of each mask and apply the colors to masks based on their centeredness locations. Therein, we can convert the entity masks to a RGB colormap.

\begin{figure}[t]
    \centering
    \includegraphics[width=\linewidth]{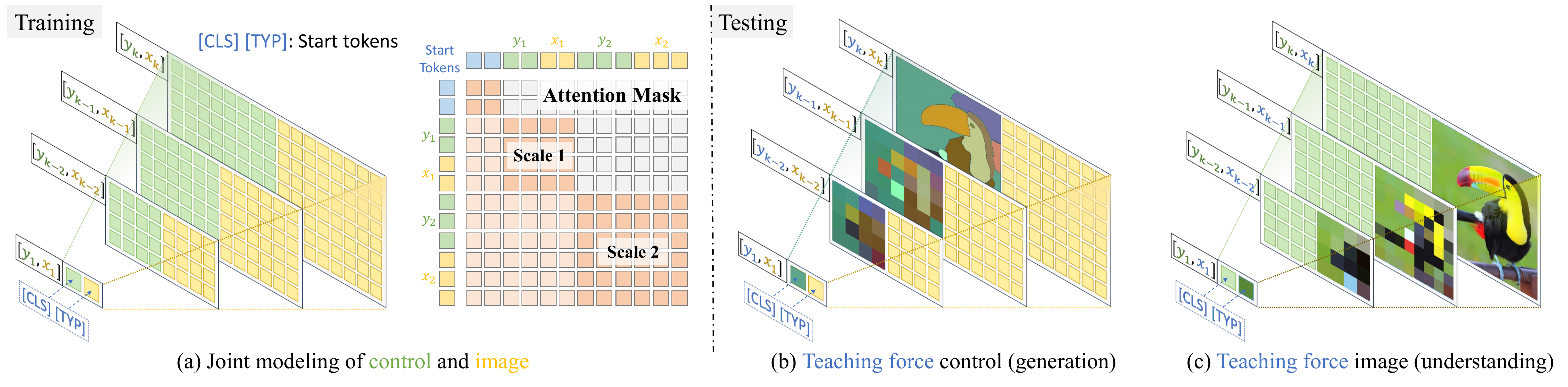}
    \caption{Illustration of ControlVAR. We jointly model the control and image during training with start tokens $[\texttt{CLS}]$ and $[\texttt{TYP}]$ to specify the semantics and control type. We conduct conditional generation by teacher forcing the AR prediction during testing.}
    \label{fig:pipeline}
\end{figure} 

\paragraph{Tokenization.} As the control and image share the same RGB representation, we can utilize the same approach to tokenize them. To represent an RGB image as a sequence of discrete tokens $(x_1,x_2,\dots,x_T)$, patch-level \cite{esser2021taming} and scale-level \cite{tian2024visual} representations have been explored. The patch-level tokenization process splits an image into $T$ patches and represents each patch as a token $x_t$ where $x_t\in[V]^{1}$ is an integer from a vocabulary of size $V$. Recently, a scale-level representation has been introduced which decomposes the image into $T$ scales where each scale is represented by a set of tokens $x_t\in[V]^{h_t\times w_t}$ (\cref{fig:teaser}(b)). $h_t\times w_t$ denotes the size of the $t$-th scale. Compared to patch-level representation, scale-level representation can better preserve the spatial locality and capture global information which are desired for conditional image generation tasks. 
This motivates us to adopt the scale-level representation in our approach. Specifically, we obtain the image tokens and control tokens using the shared tokenizer $\Phi$ as
\begin{equation}
    (x_1,x_2,\dots,x_T)=\Phi(I),\quad(y_1,y_2,\dots,y_T)=\Phi(C).
\end{equation}
Here, $x_t\in[V]^{h_t\times w_t}$ and $y_t\in[V]^{h_t\times w_t}$ share the same vocabularies, which makes it easier for joint control-image AR modeling.

\subsection{Joint Control-Image Modeling}
\label{sec:joint}
We demonstrate the network details for joint modeling in this section. Following VAR \cite{tian2024visual}, we leverage a GPT-2 style Transformer network architecture for our ControlVAR models. As shown in \cref{fig:pipeline} (a), we jointly model the control and image in each stage. A flatten operation is adopted to convert the sequence of 2D features into 1D. Full attention is enabled for both control and image tokens belonging to the same scale, which allows the model to maintain spatial locality and to exploit the global context between control and image. A standard cross entropy loss is used to supervise our autoregressive ControlVAR models.

Specifically, we employ two pre-defined special tokens $c=\texttt{[CLS]}\in[N_{cls}]^{1}$ and $c_t=\texttt{[TYP]}\in[N_{typ}]^{1}$ as the start tokens. $N_{cls}$ and $N_{tpy}$ denote the number of classes and control types respectively. $\texttt{[CLS]}$ token aims to provide semantic context for the generated image. $\texttt{[TYP]}$ token is used to select the type of pixel-level control to be generated along with the image. Following previous works \cite{chang2023muse,tian2024visual}, additional empty tokens are used to replace special tokens with a probability of $\delta$ during training to apply classifier-free guidance \cite{ho2022classifier}. 

\subsection{Sampling with Teacher-forcing Guidance.}
\label{sec:tfg}
Classifier-free guidance (CFG) \cite{ho2022classifier} was originally introduced to apply and enhance the effect of conditional controls on diffusion models without an explicit classifier. Extensive studies \cite{sanchez2023stay,chang2023muse,tian2024visual} have demonstrated that classifier-free guidance also works for AR models. 

\begin{figure}[t]
    \centering
    \includegraphics[width=\linewidth]{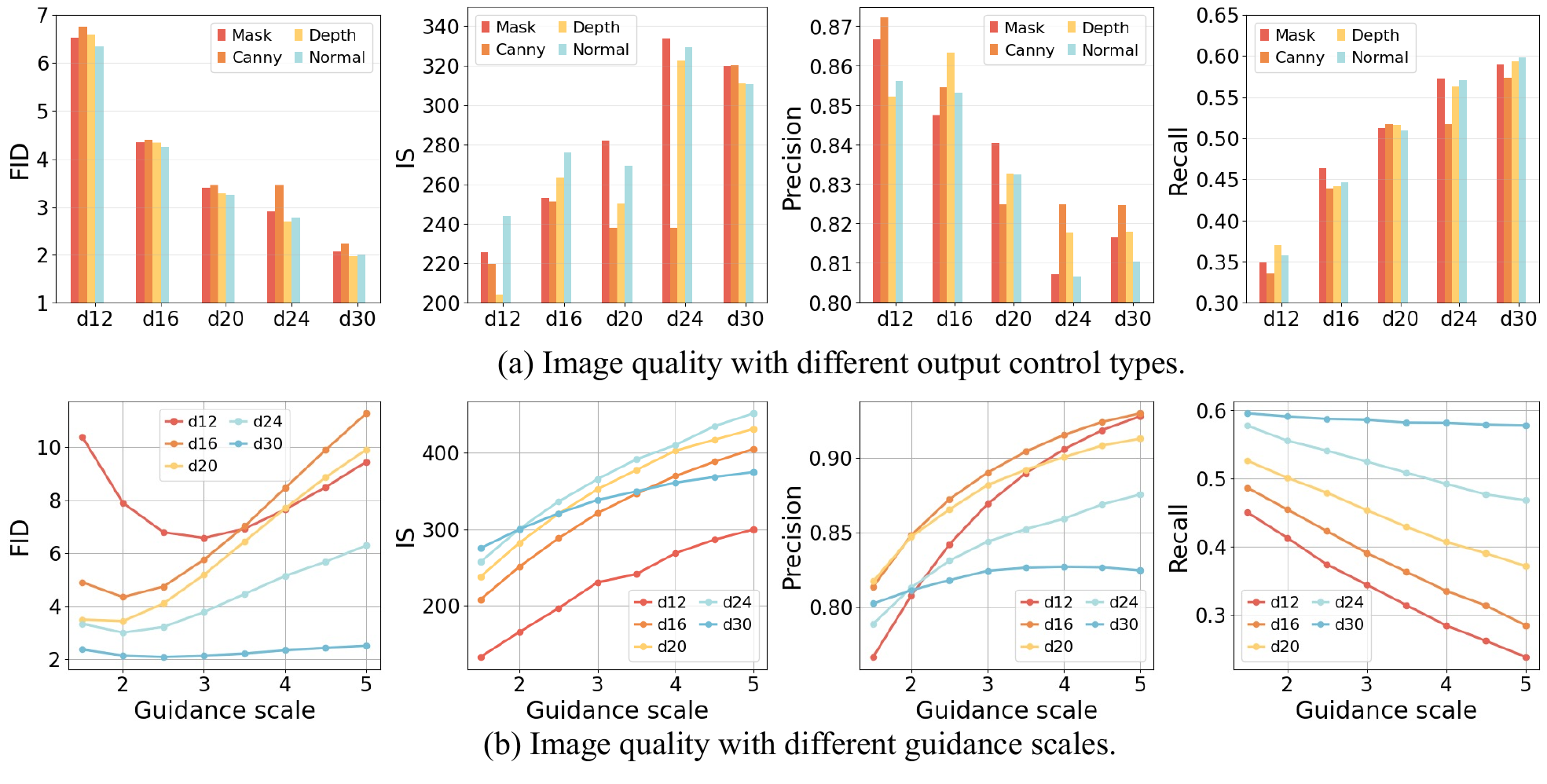}
    \caption{Joint control-image generation with (a) different output control types, (b) guidance scales.}
    \label{fig:uncond}
\end{figure}

Here, we analyze how to achieve conditional generation by using the image generation task $p(I|C,c,c_t)$ as an example. Given image $I$, pixel-level control $C$ and token-level controls $c,c_t$, CFG \cite{ho2022classifier} leverages Bayesian rule to rewrite the conditional distribution as 
\begin{equation}
\label{equ:cfg}
    p(I|C,c,c_t)\propto p(c|I,C,c_t)p(c_t|I,C)p(C|I)p(I).
\end{equation}
It can be seen that the class $c$ and control type $c_t$ are independent. By applying the Bayesian rule again, we have
\begin{equation}
\label{equ:cls}
p(c|I,C,c_t)=\frac{p(I,C|c,c_t)p(c,c_t)}{p(I,C|c_t)p(c_t)}=\frac{p(I,C|c,c_t)p(c)}{p(I,C|c_t)}.
\end{equation}
Given the AR nature of ControlVAR, 
$p(I,C|c,c_t)$ and $p(I,C|c_t)$ can be induced by using the pixel-level condition $C$ to teacher-force ControlVAR during the AR prediction. Similarly, after rewriting all terms in \cref{equ:cfg} to the form in \cref{equ:cls}, we derive an approach to sample with both pixel- and token-level controls for image generation as
\begin{equation}
\label{equ:tfg}
\begin{aligned}
x^*=x(\Rsh\!\! \emptyset|\emptyset,\emptyset)&+\gamma_{cls}(x(\Rsh\!\! C|c,c_t)-x(\Rsh\!\! C|\emptyset,c_t))\\
&+\gamma_{typ}(x(\Rsh\!\! C|\emptyset,c_t)-x(\Rsh\!\! C|\emptyset,\emptyset)) \\
&+\gamma_{pix}(x(\Rsh\!\! C|\emptyset,\emptyset)-x(\Rsh\!\! \emptyset|\emptyset,\emptyset))
\end{aligned}
\end{equation}
where $\gamma_{cls},\gamma_{typ},\gamma_{pix}$ are guidance scales for controlling the generation. As shown in \cref{fig:pipeline} (b), $x(\Rsh\!\! C|c,c_t)$ denotes the image tokens obtained by prefix $c,c_t$ and teacher forcing with $C$. $\emptyset$ denotes an empty token that avoids teacher forcing with $c,c_t$ and $C$ respectively. After obtaining the predicted tokens, the image can be decoded by a decoder as 
\begin{equation}
    I=\Phi^{-1}(x_1^*,x_2^*,\dots,x_T^*).
\end{equation} 
For the image-to-condition generation (\cref{fig:pipeline} (c)), $y^*$ can be obtained similarly by teacher forcing with $I$ and decoded similarly with the shared decoder $\Phi^{-1}$. Since 
teacher forcing is leveraged in the entire sampling process, we term the proposed strategy teacher-forcing guidance (TFG). More analysis of TFG is available in the Appendix.

\section{Experiment}

\begin{figure}[t]
    \centering
    \includegraphics[width=\linewidth]{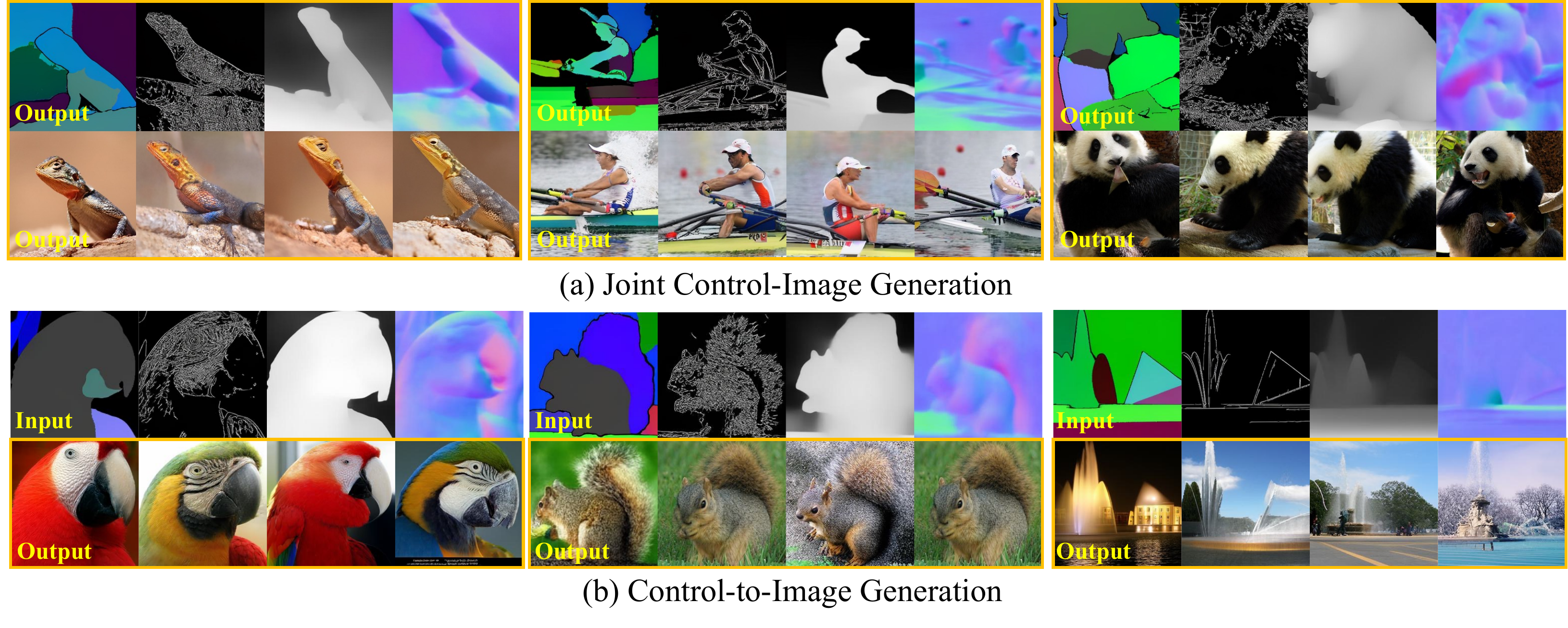}
    \caption{Visualization of (a) joint control-image generation and (b) control-to-image generation.}
    \label{fig:c_cond}
\end{figure}

\subsection{Evaluation Settings}
\paragraph{Dataset.}
We conduct all the experiments on the ImageNet \cite{deng2009imagenet} dataset. To incorporate pixel-level controls, we leverage state-of-the-art image understanding models to pseudo-label the images. Specifically, we label entity masks \cite{kirillov2023segment}, canny \cite{canny1986computational}, depth \cite{ranftl2020towards} and normal \cite{vasiljevic2019diode} for both training and validation sets. This takes 500 Tesla V100 for about 4 days. We will release the pseudo-labeled datasets to facilitate the community to further explore conditional image generation.

\paragraph{Evaluation metrics.}
We utilize Fréchet Inception Distance (FID) \cite{heusel2017gans}, Inception Score (IS) \cite{salimans2016improved}, Precision, and Recall as metrics for assessing the quality of image generation. 
However, for the image-to-control prediction where ground truth is unavailable, we rely on qualitative visualization to demonstrate the perceptual quality.

\paragraph{Implementation details.}
We follow 
VAR \cite{tian2024visual} to use a GPT-2 \cite{radford2019language} style transformer with adaptive normalization \cite{zhang2018self}. A transformer layer depth from 12 to 30 is explored. We leverage the pre-trained VAR tokenizer \cite{tian2024visual} to tokenize both image and control. We initialize the model with the weights from VAR \cite{tian2024visual} to shorten the training process. For each depth, we train the model for 30 epochs with an Adam optimizer. We follow the same learning rate and weight decay as VAR. During training, we sample each control type uniformly. To apply the classifier-free guidance, we replace class and control type conditions with empty tokens with 0.1 probability. We train the model with $\textrm{batchsize}=128$ for all the experiments. During inference, we utilize top-$k$ top-$p$ sampling with $k=900$ and $p=0.96$. We utilize $256\times 256$ image size for all experiments. For simplicity, we leverage $\gamma_{cls}=\gamma_{typ}=\gamma_{pix}$ for all the experiments.

\subsection{Performance Analysis}

\paragraph{Joint image-control generation.}
We demonstrate the performance of ControlVAR with different output control types, model sizes and guidance scales as shown in \cref{fig:uncond} (a) and \cref{fig:uncond} (b). As the model size increases, we notice ControlVAR performs better generation capability accordingly. Among all control types, jointly generating canny and image leads to a slightly inferior performance compared to other types. We consider the complex pattern of canny may impose difficulty in generating corresponding images thus leading to the degradation. In addition, we notice the optimum FID can be achieved with a guidance scale between 2 to 3. Though further increasing the guidance scale can still improve the IS, it will limit the mode diversity. We demonstrate qualitative visualization of joint generation in \cref{fig:c_cond} (a) which shows high-quality and aligned image-control pairs.
\begin{wraptable}{r}{0.45\textwidth}   
    \centering
    \vspace{-0.3cm}
    \scalebox{1}{
    \hspace{-0.4cm}
    \begin{tabular}{c|c|c|c|c} 
        Depth & 16 & 20 & 24 & 30 \\
        \hline
        VAR & 3.60 & 2.95 & 2.33 & 1.97 \\
        ControlVAR & 4.25 & 3.25 & 2.69 & 1.98\\
    \end{tabular}}
    \vspace{-0.3cm}
    \caption{Image FID compared to VAR.}
    \vspace{-0.4cm}
    \label{tab:comp_var}
\end{wraptable}
Furthermore, we compare the image FID with pure image generation model VAR \cite{tian2024visual} in \cref{tab:comp_var}. We notice that ControlVAR shows a slight performance degradation compared to VAR which can be due to the difficulty enrolled to incorporate additional controls. As the model size increases, we notice the performance gap shrinks, indicating joint modeling of image and control may require more network capacity compared to image-only modeling.

\begin{figure}[t]
    \centering
    \includegraphics[width=\linewidth]{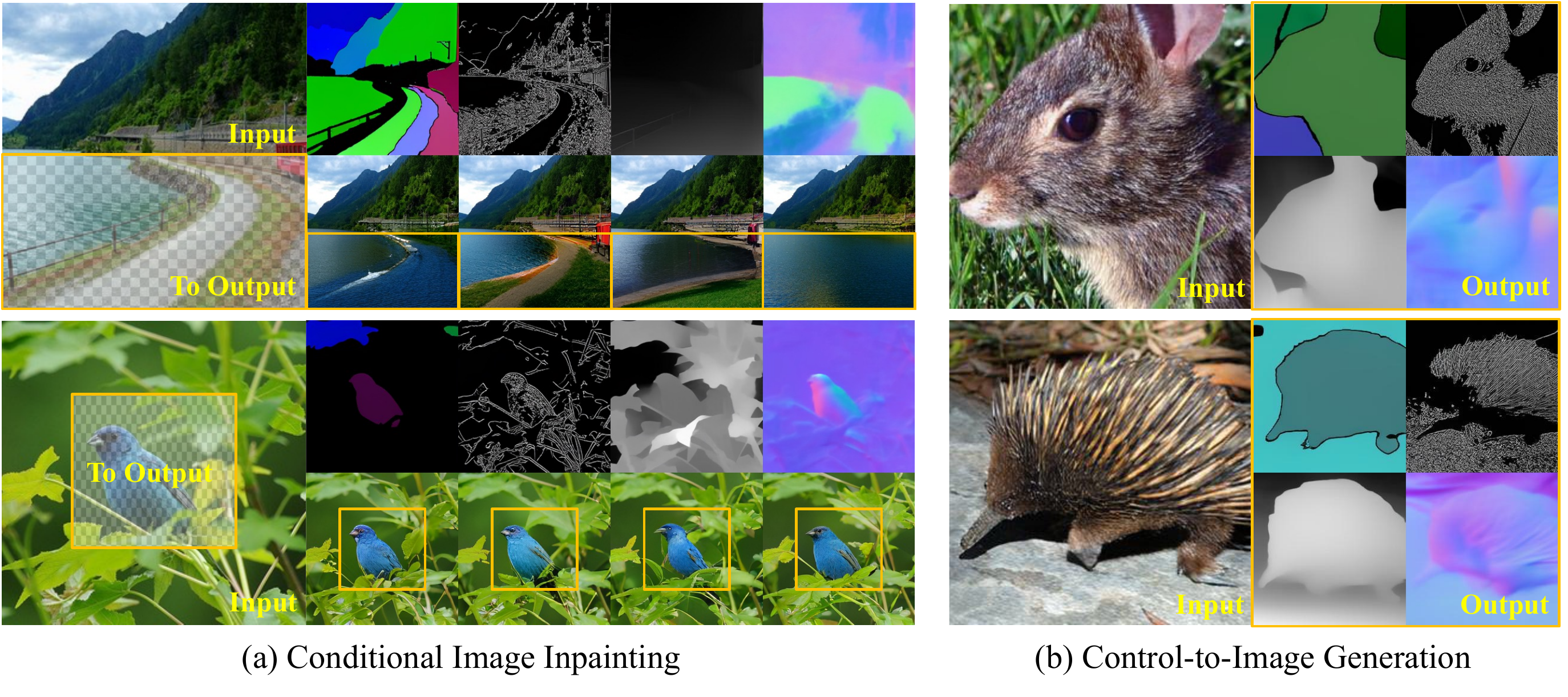}
    \caption{Visualization of conditional image inpainting (given pixel-level control and partial image).}
    \label{fig:inpaint}
\end{figure}

\begin{figure}[t]
    \centering
    \includegraphics[width=\linewidth]{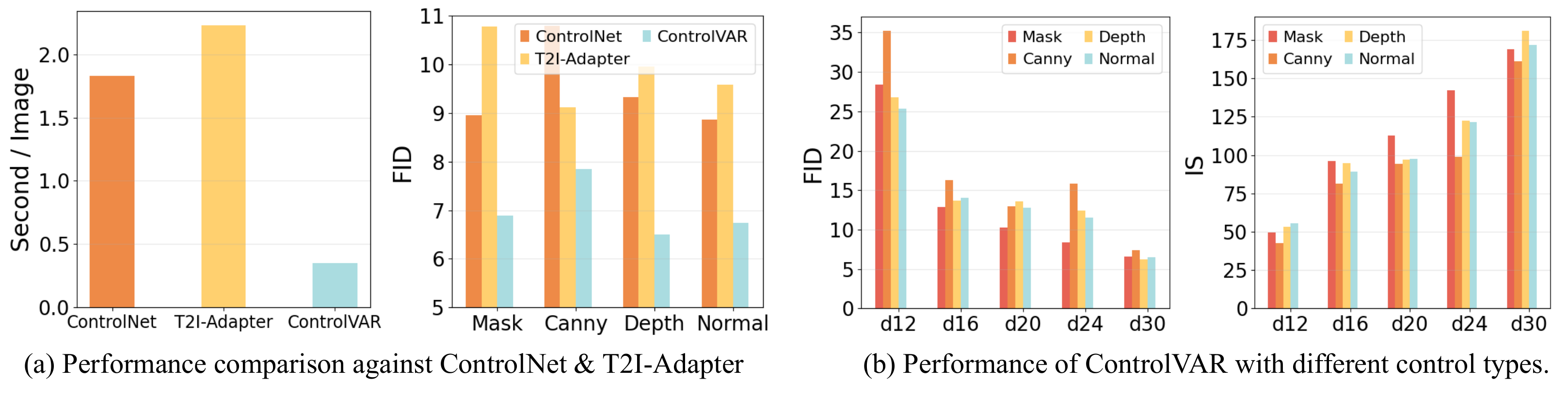}
    \vspace{-0.6cm}
    \caption{Quantitative results of conditional image generation.}
    \vspace{-0.1in}
    \label{fig:cond_analysis}
\end{figure}
\begin{figure}[t]
    \centering
    \includegraphics[width=\linewidth]{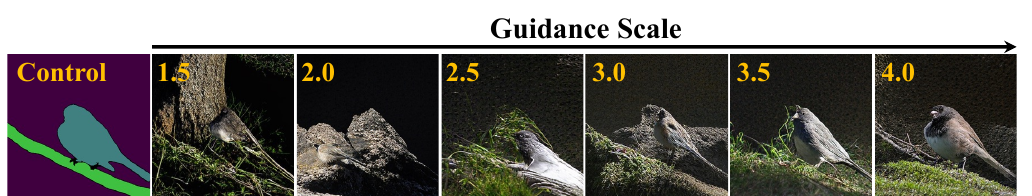}
    \caption{Visualization of images generated with different guidance scales.}
    \label{fig:tfg}
\end{figure}

\paragraph{Conditional image generation.}

We introduce two baseline methods - ControlNet \cite{zhang2023adding} and T2I-Adapter \cite{mou2023t2i} to compare the conditional generation capability. We train both baselines on the same datasets as ours with the Diffuser \cite{von-platen-etal-2022-diffusers} implementation (for a fair comparison, ImageNet-pretrained LDM \cite{rombach2021highresolution} is used as the base model). We compare the image generation quality in terms of FID and speed in \cref{fig:cond_analysis}. We notice that ControlVAR achieves obvious superior FID compared to baselines. We evaluate the inference speed with $\mathrm{batchsize}=1$ on a single H100 GPU.
We notice that ControlVAR inference is at least 5 times faster than the compared methods. We further explore the generation capability with different model sizes as shown in \cref{fig:cond_analysis} (b), we notice that the model's generation capability keeps improving as the model size increases. Similar to the joint image-control generation, we notice that canny-conditioned generation shows an inferior performance due to its complex pattern.

\begin{wraptable}{r}{0.5\textwidth}   
    \centering
    \vspace{-0.2cm}
    \scalebox{0.95}{
    \hspace{-0.4cm}
    \begin{tabular}{c|c|c|c} 
        Method & VQ-GAN & IQ-VAE & ControlVAR \\
        \hline
        FID & 35.5 & 29.77 & 9.72 \\
    \end{tabular}}
    \vspace{-0.3cm}
    \caption{FID comparison on ADE20K.}
    \vspace{-0.4cm}
    \label{tab:ade}
\end{wraptable}
We further compare ControlVAR with conditional AR models - VQ-GAN \cite{esser2021taming} and IQ-VAE \cite{zhan2022auto}. We finetune ControlVAR on ADE 20K for 1 epoch and report the FID of the generated images in \cref{tab:ade}. ControlVAR demonstrates superior performance compared to previous AR methods.

\paragraph{Conditional image inpainting.}
ControlVAR can support more complex image generation tasks by teacher-forcing with partial image/control. As shown in \cref{fig:inpaint} (a), we showcase the conditional image inpainting results where pixel-level control and partial image are given to complete the missing part of the image. We notice that the contents align well with both the given control and image.

\paragraph{Image-to-control prediction.}
ControlVAR is also capable of image understanding tasks by teacher-forcing with images during inference. As shown in \cref{fig:inpaint} (b), we demonstrate the visualization of the generated controls given images. Since the pseudo labels that we use during training and inference are mediocre in quality, we do not focus on the understanding capability of ControlVAR in this paper and leave it for future work instead.

\subsection{Ablation Experiments}

\paragraph{Module effectiveness.}
We conduct ablation experiments to validate the effectiveness of components in ControlVAR. We start with a depth 16 baseline which models the control and image in different scales without joint modeling. 
\cref{tab:ablation} shows the impact of adding each component. We notice an obvious performance improvement by using joint modeling. Unlike the baseline setting, joint modeling enables both control and image to interact with each other on the same scale leading to better pixel-level alignment for the teacher forcing during inference. In addition, with the multi-control training and teacher forcing guidance, ControlVAR achieves 5.19 and 15.21 FID for joint control-image and control-to-image generation respectively. During inference, we linearly anneal the guidance scale using $\gamma\cdot\frac{t}{T}$ (where $t$ is the iteration number, $T$ is the total AR iterations, and $\gamma$ is a constant hyperparameter) which brings another 0.84 and 2.24 FID gains. Lastly, by scaling the model size to depth 30, we achieve the best results of 2.09 and 6.57 FID.

\begin{table*}
\centering
\scalebox{1}{
\begin{tabular}{
c|lp{1.7cm}
<{}p{1.7cm}<{}|p{1.7cm}<{}p{1.7cm}<{}} 
\hline
\multirow{2}*{ID} & \multirow{2}*{Method} & \multicolumn{2}{c}{Joint Control-Image} & \multicolumn{2}{c}{Control-to-Image}\\
\cline{3-6}
~ & ~ & FID$\downarrow$ & IS$\uparrow$ & FID$\downarrow$ & IS$\uparrow$\\
\hline
\rowcolor{gray!10}1 & Baseline (w/o joint modeling) & 12.23 & 119.65 & 35.92 & 42.50 \\
\hline
2 & + Joint modeling & 9.74\color{fgreen}{$_{-2.49}$}  & 142.08\color{fgreen}{$_{+22.43}$} & 17.44\color{fgreen}{$_{-18.48}$} & 77.38\color{fgreen}{$_{+34.88}$}  \\
3 & + Multi-control training & 5.19\color{fgreen}{$_{-4.55}$} & 223.10\color{fgreen}{$_{+81.02}$} & 16.33\color{fgreen}{$_{-1.11}$} & 98.62\color{fgreen}{$_{+21.24}$} \\
4 & + Teacher-forcing guidance & - & - & 15.21\color{fgreen}{$_{-1.12}$} & 95.44 \color{fgreen}{$_{+3.18}$} \\
5 & + Guidance scaling  & 4.35\color{fgreen}{$_{-0.84}$}  & 253.08\color{fgreen}{$_{+29.98}$} & 12.97\color{fgreen}{$_{-2.24}$} & 96.42\color{fgreen}{$_{+0.98}$} \\
6 & + Larger model size  & 2.09\color{fgreen}{$_{-2.26}$} & 337.86\color{fgreen}{$_{+84.78}$} & 6.57\color{fgreen}{$_{-6.40}$} & 173.02\color{fgreen}{$_{+76.6}$} \\
\hline
\end{tabular}
}
\label{tab:ablation}
\caption{Ablation study on components in ControlVAR. We evaluate the FID and IS on the ImageNet validation set with masks as the target controls.}
\label{tab:ablation}
\end{table*}

\paragraph{Teacher forcing guidance.} Given the same mask control, we further visualize the images generated with different guidance scales in \cref{fig:tfg}. As the guidance scale increases, the generated contents align more with the given control, indicating that the TFG can effectively enhance the guidance effect.

\paragraph{Generalization to unseen tasks.}
\begin{wrapfigure}{r}{0.28\textwidth}
  \centering
  \includegraphics[width=0.28\textwidth]{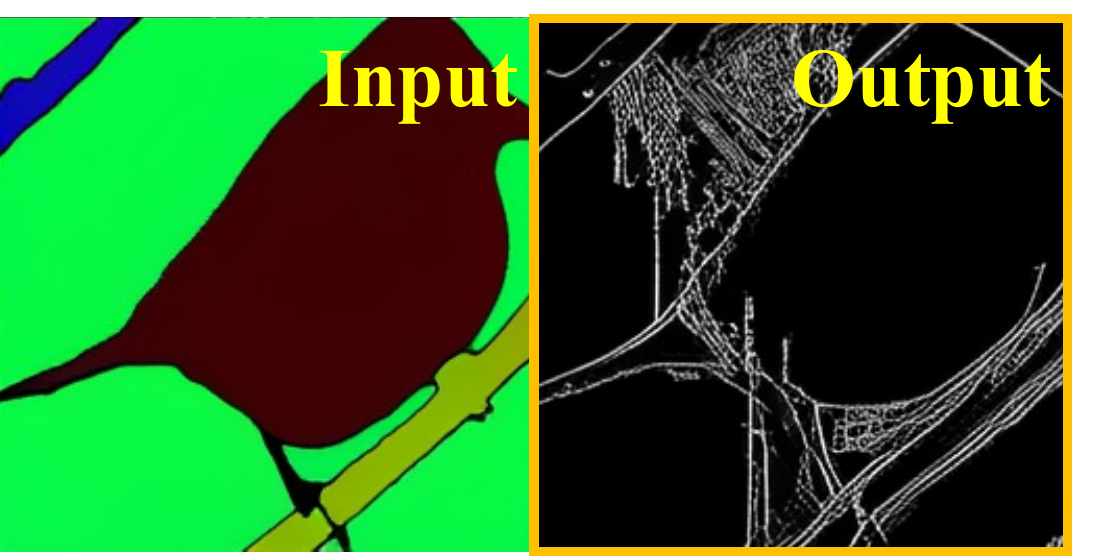} 
  \caption{Mask-to-Canny.}
  \label{fig:zeroshot}
\end{wrapfigure}
As shown in \cref{fig:zeroshot}, we conduct an unseen task by teacher-forcing a mask in the AR prediction and setting the type token to predict the canny. We notice ControlVAR can successfully generate aligned results. We optimize ControlVAR with the joint distribution between the image and controls $\sum_n p(I,C_n)$ during training which can be assumed as an alternating optimization of $p(I,\{C_n\})$. We consider this to explain the observed zero-shot capability with unseen control-to-control tasks. More visualizations are available in the Appendix.
\section{Conclusion}
In this paper, we present ControlVAR, an autoregressive (AR) approach for conditional generation. 
Unlike traditional conditional generation models that leverage prefix pixel-level controls, \eg, mask, canny, normal, and depth, ControlVAR jointly models image and control conditions during training and enables flexible conditional generation during testing by teacher forcing.  
Inspired by the classifier-free guidance, we introduce a teacher-forcing guidance strategy to facilitate controllable sampling. Comprehensive and systematic experiments are conducted to demonstrate the effectiveness and characteristics of ControlVAR, showcasing its superiority over powerful DMs in handling multiple conditions for diverse conditional generation tasks.

\paragraph{Limitations.}
In spite of ControlVAR's high performance on image generation with heterogeneous pixel-level controls, it does not support text prompts and therefore cannot be directly leveraged with natural language guidance. Developing text-guided capability can be achieved by replacing the class token with the language token, \eg, CLIP token \cite{radford2021learning}, which is left as our future focus.

\bibliography{iclr2025_conference}
\bibliographystyle{iclr2025_conference}

\renewcommand{\thetable}{{\Alph{table}}}
\renewcommand{\thefigure}{{\Alph{figure}}}
\renewcommand{\thesection}{{\Alph{section}}}
\setcounter{figure}{0}   
\setcounter{table}{0}   
\setcounter{section}{0} 

\clearpage
\appendix
\begin{figure}[t]
    \centering
    \includegraphics[width=\linewidth]{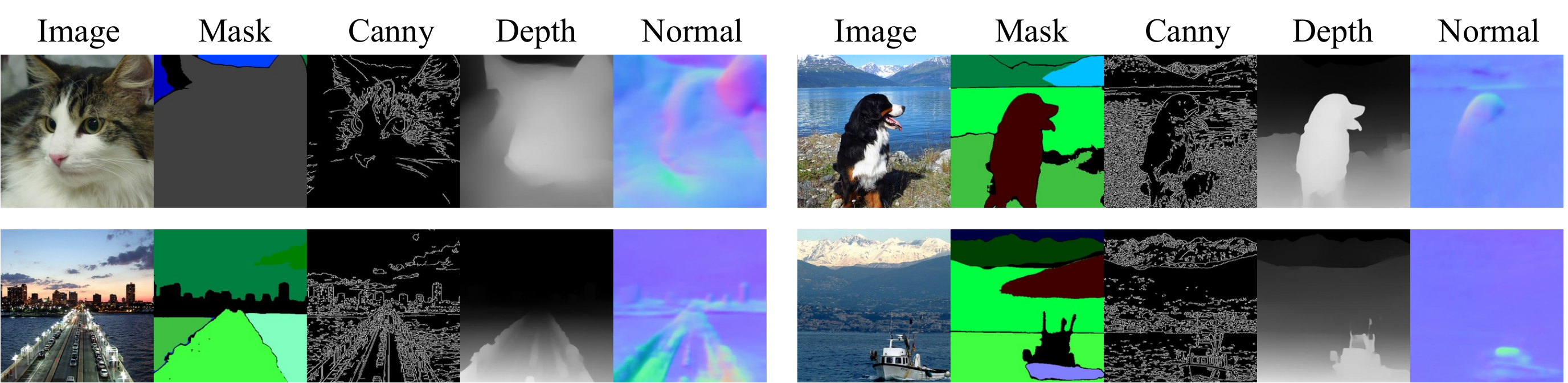}
    \caption{Example of image and corresponding controls in the pseudo-labeled dataset.}
    \label{fig:visualization_dataset}
\end{figure}

\section{Dataset Creation}

\begin{wraptable}{r}{0.6\textwidth}   
    \centering
    \vspace{-0.3cm}
    \scalebox{1}{
    \hspace{-0.4cm}
    \begin{tabular}{c|c|c|c|c} 
        Type & Mask & Canny & Depth & Normal \\
        \hline
        \# Sample & 1277548 & 1277653 & 1277636 & 1277639 \\
    \end{tabular}}
    \vspace{-0.2cm}
    \caption{Statistics of the generated dataset.}
    \vspace{-0.4cm}
    \label{tab:data}
\end{wraptable} 
We conduct all the experiments on the ImageNet \cite{deng2009imagenet} dataset. To incorporate pixel-level controls, we leverage state-of-the-art image understanding models to pseudo-label the images. Specifically, we label entity masks \cite{kirillov2023segment}, canny \cite{canny1986computational}, depth \cite{ranftl2020towards} and normal \cite{vasiljevic2019diode} for both training and validation sets. This takes 500 Tesla V100 for about 4 days. 
We demonstrate the label number after filtering in \cref{tab:data}. In addition, we also manually check the quality of the pseudo labels. We show a visualization of the generated datasets in \cref{fig:visualization_dataset}. We notice that image understanding models predict reasonable results on ImageNet images.

\section{Discussion of the Teacher Forcing Guidance}
Inspired by the classifier-free guidance \cite{ho2022classifier} from diffusion models, we empirically find a similar form of guidance that can be used for autoregressive sample conditional images based on teacher forcing. In this section, we aim to analyze the spirit of classifier-free guidance (CFG) and analogy it to our teacher-forcing guidance (TFG).

\subsection{Classifier-free Guidance}
For the image generation task $p(I|C,c,c_t)$, given image $I$, pixel-level control $C$ and token-level controls $c_{c},c_t$, CFG leverages Bayesian rule to rewrite the conditional distribution as 
\begin{gather*}
\label{equ:gd_cfg}
    p(I|C,c,c_t)=\frac{p(c|I,C,c_t)p(c_t|I,C)p(C|I)p(I)}{p(C,c,c_t)} \\
    \implies \log p(I|C,c,c_t)=\log p(c|I,C,c_t)+\log p(c_t|I,C)+\log p(C|I)+\log p(I) - \log p(C,c,c_t) \\
    \implies \nabla_I\log p(I|C,c,c_t)=\nabla_I\log p(c|I,C,c_t)+\nabla_I\log p(c_t|I,C)+\nabla_I\log p(C|I)+\nabla_I\log p(I) 
\end{gather*}
By applying the Bayesian rule again, we have
\begin{gather*}
\label{equ:cls_supp}
p(c|I,C,c_t)=\frac{p(I|c,c_t,C)p(c,c_t,C)}{p(I|c_t,C)p(c_t,C)} \\
\implies \nabla_I \log p(c|I,C,c_t)=\nabla_I\log p(I|c,c_t,C) - \nabla_I\log p(I|c_t,C).
\end{gather*}
Similarly, by applying the Bayesian rule to all terms, we have
\begin{equation*}
\begin{aligned}
    \nabla_I\log p(I|c,c_t,C)=&\nabla_I\log p(I) \\
    +&\nabla_I \log p(I|c,c_t,C) - \nabla_I \log p(I|c_t,C) \\
    +&\nabla_I \log p(I|c_t,C) - \nabla_I \log p(I|C) \\
    +&\nabla_I \log p(I|C) - \nabla_I \log p(I) \\
\end{aligned}
\end{equation*}
In the diffusion models, $\nabla_I\log p(I|*)$ is represented by the logits outputted by the diffusion-UNet. In this way, during inference, the classifier-free guidance can be calculated as
\begin{equation*}
\begin{aligned}
    x^*=&x(\emptyset,\emptyset,\emptyset) \\
    +&\gamma_c(x(c,c_t,C) - x(\emptyset, c_t,C)) \\
    +&\gamma_{c_t}(x(\emptyset,c_t,C) - x(\emptyset,\emptyset,C)) \\
    +&\gamma_{C}(x(\emptyset,\emptyset,C) - x(\emptyset,\emptyset,\emptyset)) \\
\end{aligned}
\end{equation*}
where $\gamma_C,\gamma_c,\gamma_{c_t}$ are the guidance scales that are used to adjust the amplitude to apply the conditional guidance. $\emptyset$ denotes leveraging a special empty token to replace the original token to disable the additional conditional information \cite{ho2022classifier}.

\subsection{Teacher Forcing Guidance}
Classifier-free guidance has been proven to be effective in AR models \cite{chang2023muse,tian2024visual} which take the same form as diffusion models as 
\begin{gather*}
    p(I|C,c,c_t)\propto p(c|I,C,c_t)p(c_t|I,C)p(C|I)p(I).
\end{gather*}
In ControlVAR, we model the joint distribution of the controls and images. Therefore, we leverage a different extension of the probabilities as
\begin{equation*}
    p(c|I,C,c_t)=\frac{p(I,C|c,c_t)p(c,c_t)}{p(I,C|c_t)p(c_t)}=\frac{p(I,C|c,c_t)p(c)}{p(I,C|c_t)}
\end{equation*}
where $p(I,C|c,c_t)$ and $p(I,C|c,c_t)$ can be found from the output of ControlVAR. We follow previous works \cite{tian2024visual,esser2021taming} to ignore the constant probabilities $p(c)$. By rewriting all terms with Baysian rule, we have
\begin{equation*}
\begin{aligned}
    \log p(I|C,c,c_t) \propto &\log p(I) \\
    + &\log p(I,C|c,c_t) -  \log p(I,C|c_t) \\
    + &\log p(I,C|c_t) - \log p(I,C) \\
    + &\log p(I,C) - \log p(I). 
\end{aligned}
\end{equation*}
This corresponds to the image logits as discussed in the \cref{equ:tfg}
\begin{equation}
\begin{aligned}
x^*=x(\Rsh\!\! \emptyset|\emptyset,\emptyset)&+\gamma_{cls}(x(\Rsh\!\! C|c,c_t)-x(\Rsh\!\! C|\emptyset,c_t))\\
&+\gamma_{typ}(x(\Rsh\!\! C|\emptyset,c_t)-x(\Rsh\!\! C|\emptyset,\emptyset)) \\
&+\gamma_{pix}(x(\Rsh\!\! C|\emptyset,\emptyset)-x(\Rsh\!\! \emptyset|\emptyset,\emptyset))
\end{aligned}
\end{equation}
where $\gamma_{cls},\gamma_{typ},\gamma_{pix}$ are guidance scales for controlling the generation.

\section{Full Results of Performance Analysis}

\subsection{Details of Evaluation Metrics}
\label{sec:appendix_pretrain-metric}
\textbf{Fr\'{e}chet Inception Distance (FID)} \cite{heusel2017gans}. 
FID measures the distance between real and generated images in the feature space of an ImageNet-1K pre-trained classifier \cite{szegedy2016rethinking}, indicating the similarity and fidelity of the generated images to real images.


\textbf{Inception Score (IS)} \cite{salimans2016improved}. 
IS also measures the fidelity and diversity of generated images. 
It consists of two parts: the first part measures whether each image belongs confidently to a single class of an ImageNet-1K pre-trained image classifier \cite{szegedy2016rethinking} and the second part measures how well the generated images capture diverse classes.

\textbf{Precision and Recall} \cite{kynkaanniemi2019improved}. 
The real and generated images are first converted to non-parametric representations of the manifolds using k-nearest neighbors, on which the Precision and Recall can be computed.
Precision is the probability that a randomly generated image from estimated generated data manifolds falls within the support of the manifolds of estimated real data distribution.
Recall is the probability that a random real image falls within the support of generated data manifolds. 
Thus, precision measures the general quality and fidelity of the generated images, and recall measures the coverage and diversity of the generated images. 

\subsection{Joint Control-Image Generation}
\begin{figure}[t]
    \centering
    \includegraphics[width=\linewidth]{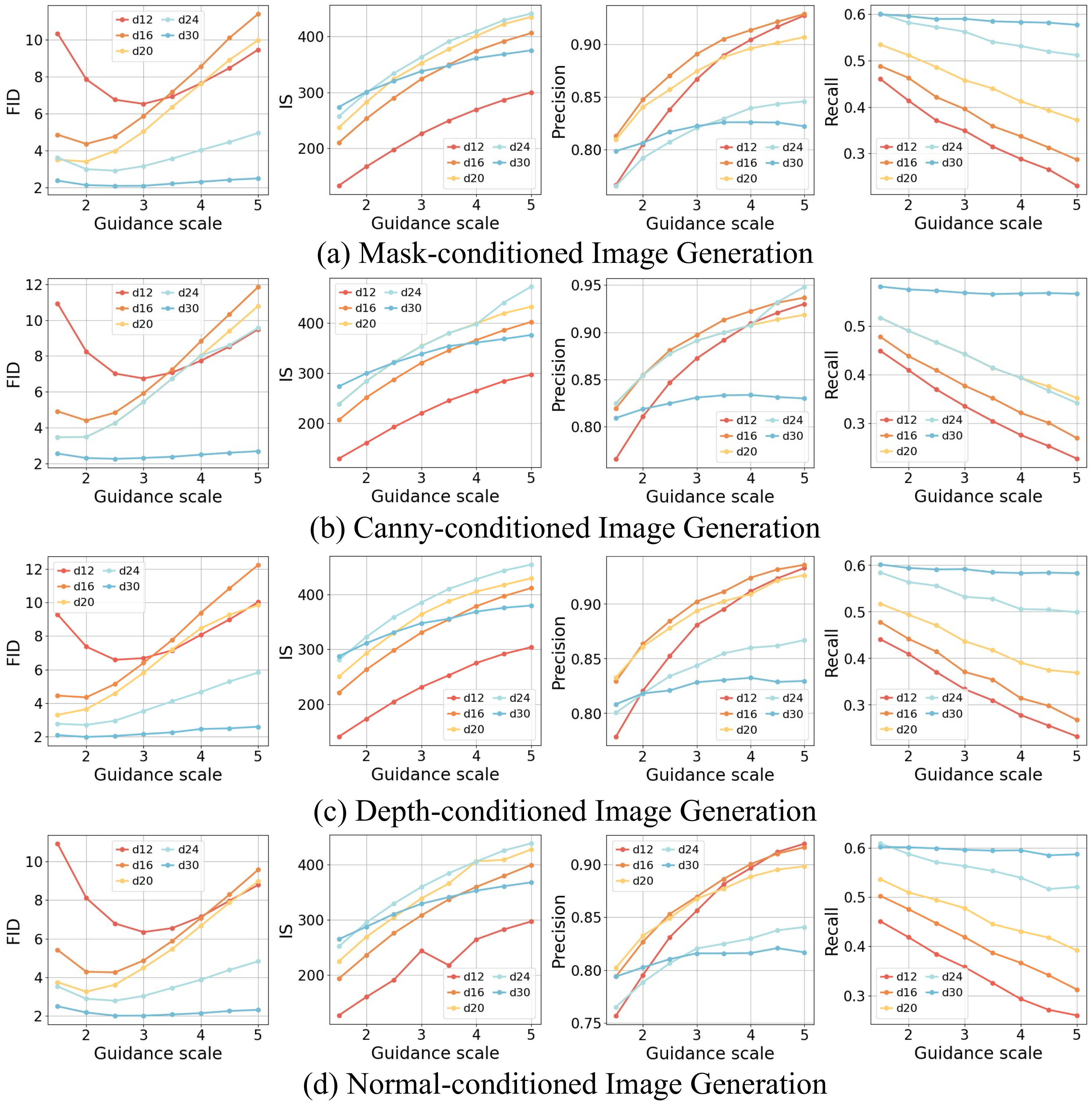}
    \caption{Performance of joint image-control generation for different control types. The performance is evaluated on the ImageNet validation set with our created pseudo labels.}
    \label{fig:uncond_supp}
\end{figure}

We demonstrate more detailed results for each control type for joint control-image generation in \cref{fig:uncond_supp}. As the model size increases, ControlVAR demonstrates better performance. The control generated along with the image shows a minor impact on the image quality.

\subsection{Control-to-Image Generation}
\begin{figure}[t]
    \centering
    \includegraphics[width=\linewidth]{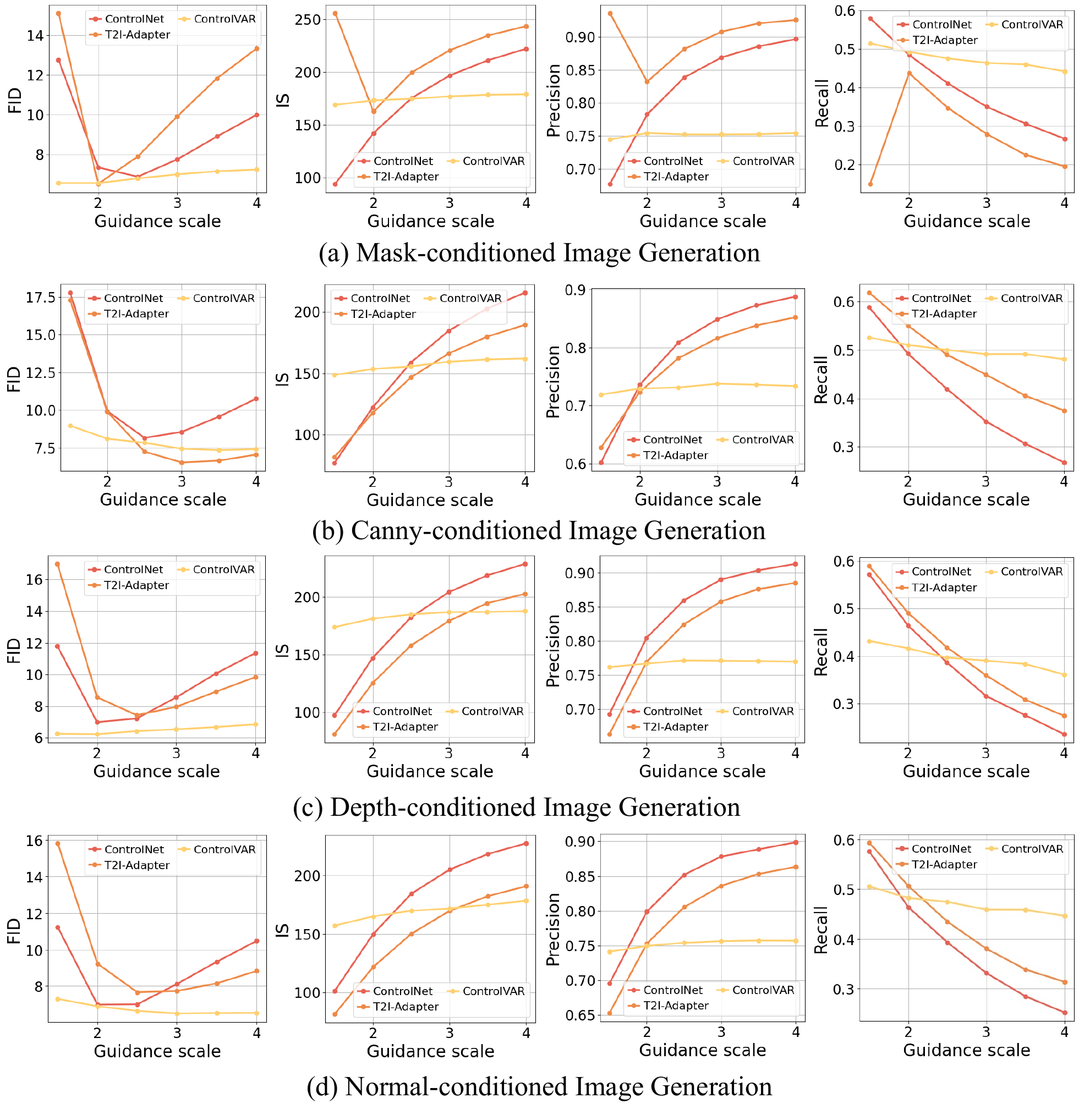}
    \caption{Performance of conditional generation for different condition types. We first introduce two baseline methods - ControlNet \cite{zhang2023adding} and T2I-Adapter \cite{mou2023t2i} to compare the conditional generation capability. We train both baselines on the same datasets as ours with the Diffuser \cite{von-platen-etal-2022-diffusers} implementation (for a fair comparison, ImageNet-pretrained LDM \cite{rombach2021highresolution} is used as the base model). The performance is evaluated on the ImageNet validation set with our created pseudo labels.}
    \label{fig:cond_cfg_comparison}
\end{figure}

We demonstrate more results when comparing with baseline models - ControlNet and T2I-Adapter in \cref{fig:cond_cfg_comparison} and \cref{fig:cgen_supp}. The performance is evaluated on the ImageNet validation set with our created pseudo labels. We notice that ControlVAR demonstrates a superior performance for different tasks. Specifically, we notice that ControlNet outperforms ControlVAR for canny-conditioned image generation. We consider this to be due to the difficulty of handling the joint modeling of the complex canny and image. In addition, we notice that when the guidance scale increases, ControlNet and T2I-Adapter demonstrate a superior inception score and an inferior FID, we consider this attributed to the increasing mode collapse resulting from the larger guidance scale. In contrast, the performance of ControlVAR is more robust.

\begin{figure}[h]
    \centering
    \includegraphics[width=\linewidth]{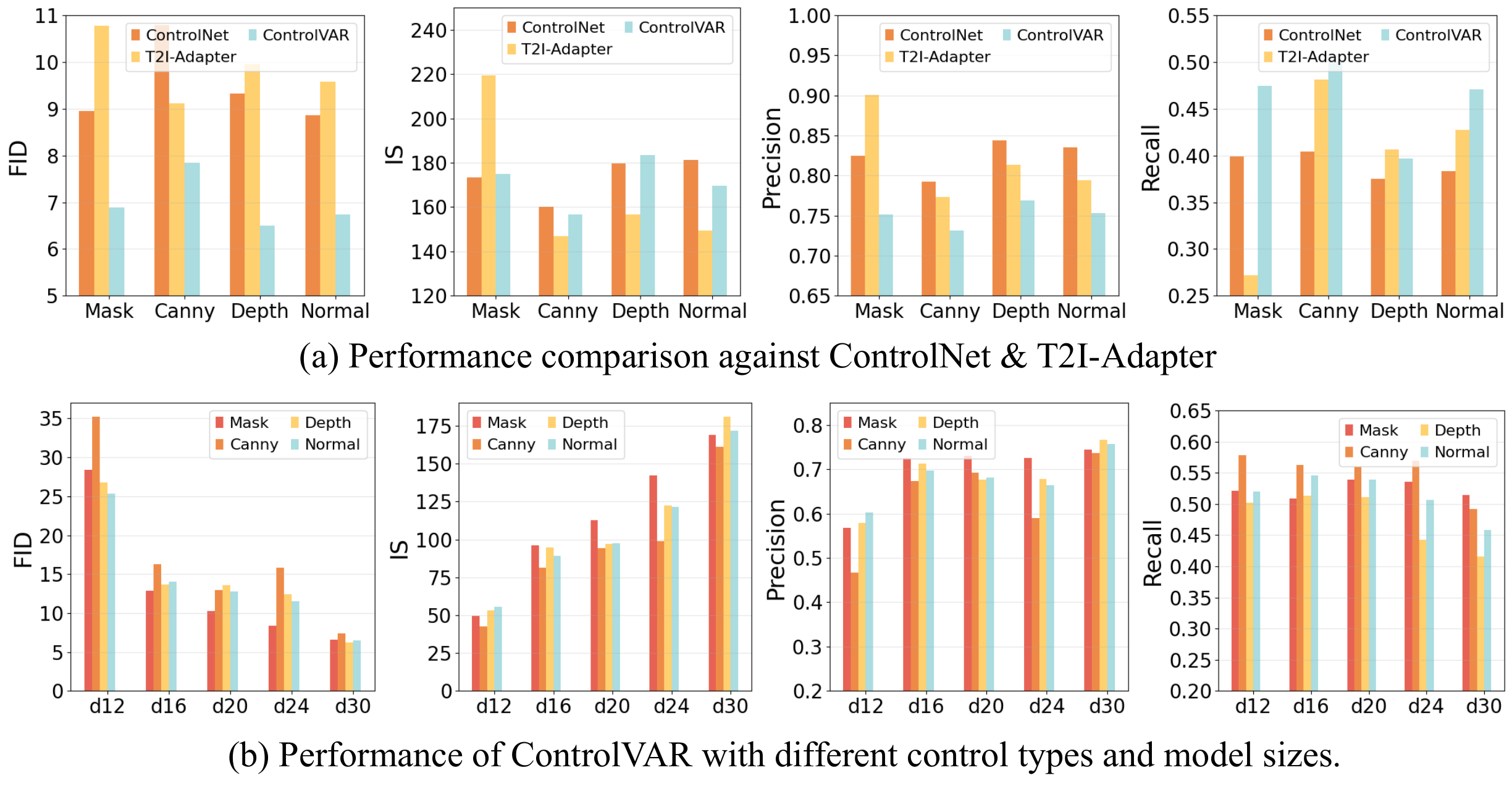}
    \caption{Performance of conditional generation for different condition types. We first introduce two baseline methods - ControlNet \cite{zhang2023adding} and T2I-Adapter \cite{mou2023t2i} to compare the conditional generation capability. We train both baselines on the same datasets as ours with the Diffuser \cite{von-platen-etal-2022-diffusers} implementation (for a fair comparison, ImageNet-pretrained LDM \cite{rombach2021highresolution} is used as the base model). The performance is evaluated on the ImageNet validation set with our created pseudo labels.}
    \label{fig:cgen_supp}
\end{figure}

\begin{figure}[t]
    \centering
    \includegraphics[width=\linewidth]{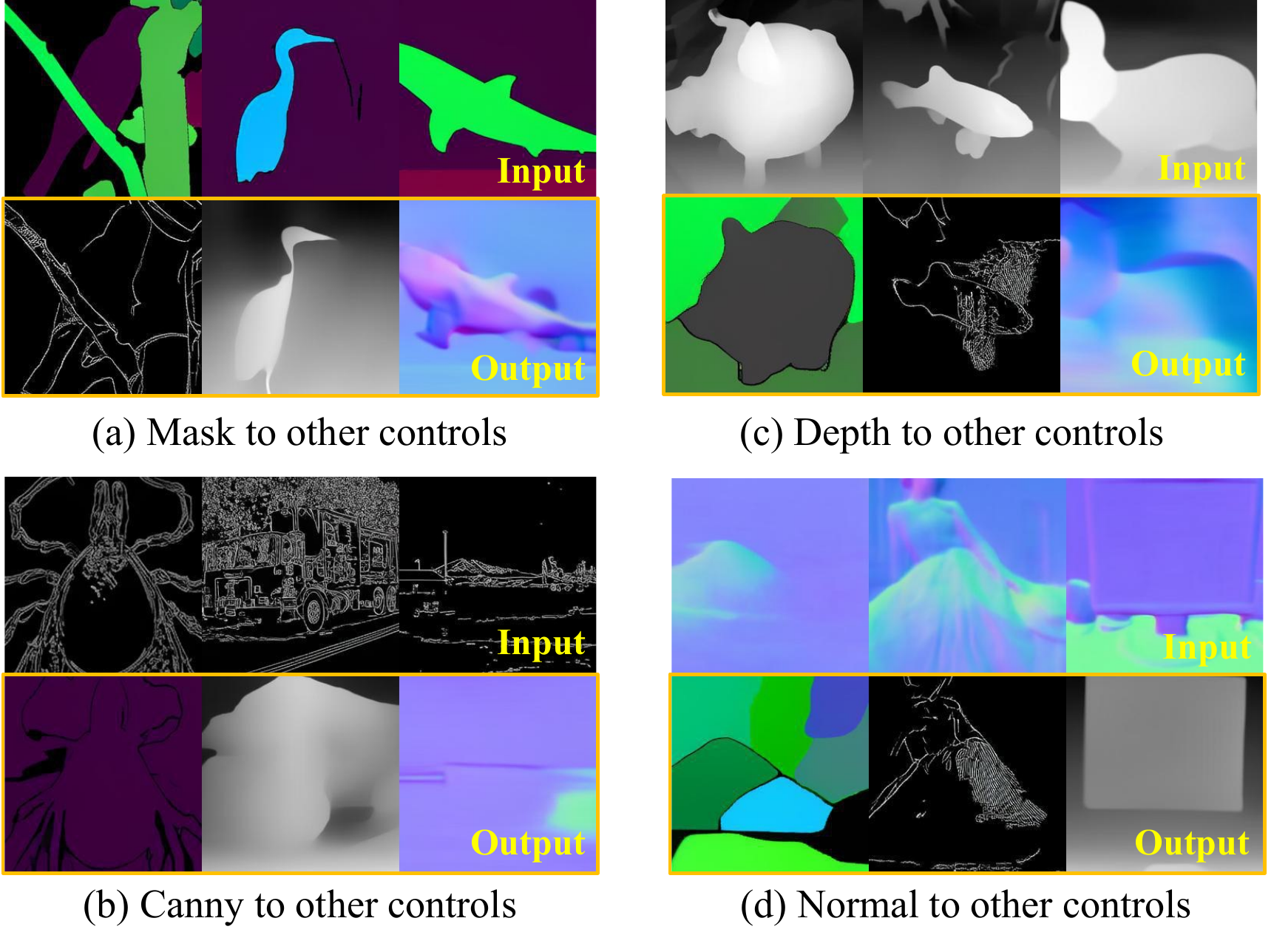}
    \caption{Qualitative visualization for zero-shot condition understanding task. The {\color{yellow}{yellow}} boxes denote the predicted controls.}
    \label{fig:zero_supp}
\end{figure}


\section{More visualization}
\begin{figure}[t]
    \centering
    \includegraphics[width=0.95\linewidth]{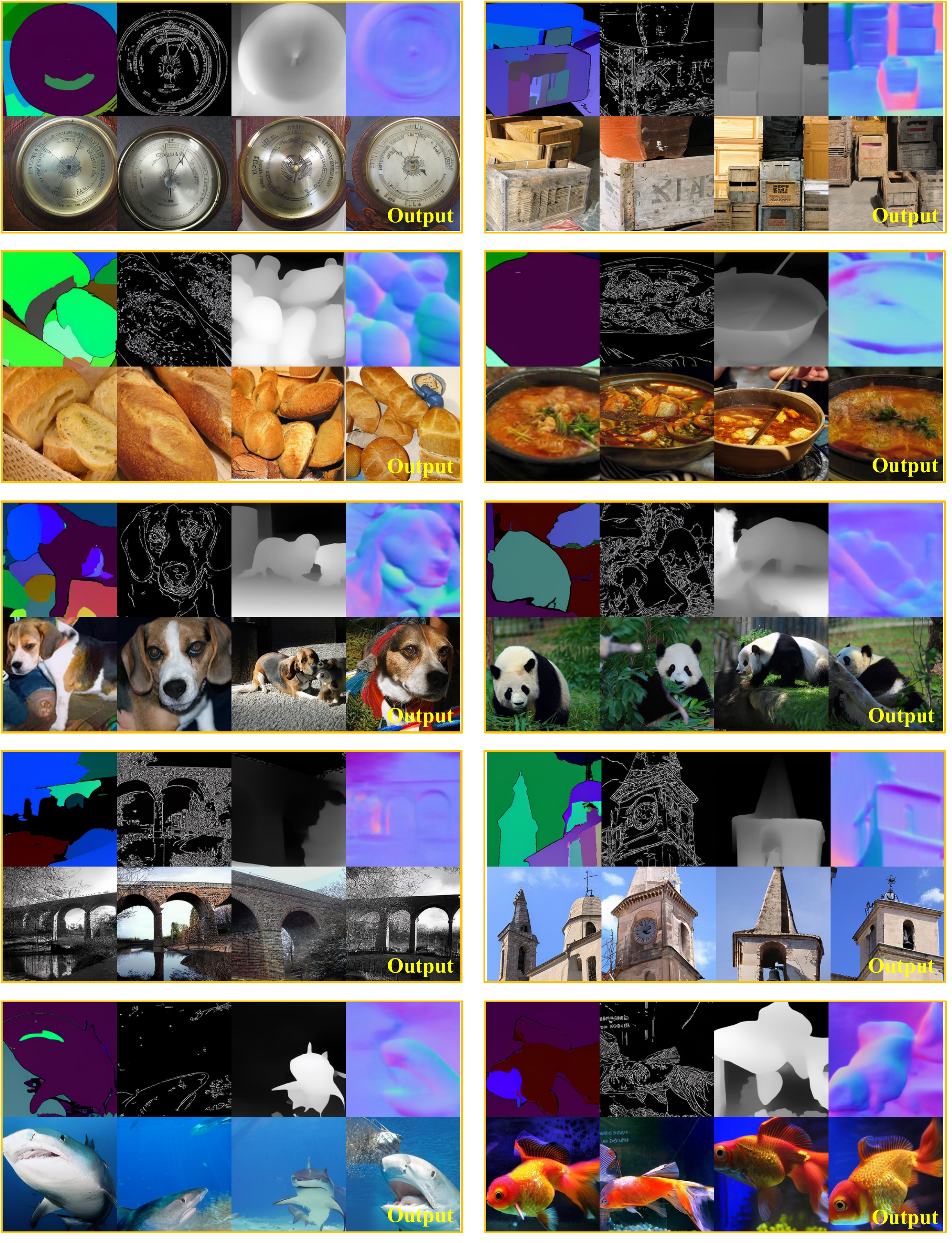}
    \caption{Qualitative visualization for joint control-image generation task. The {\color{yellow}{yellow}} boxes denote the predicted images \& controls.}
    \label{fig:joint_supp}
\end{figure}

\begin{figure}[t]
    \centering
    \includegraphics[width=0.9\linewidth]{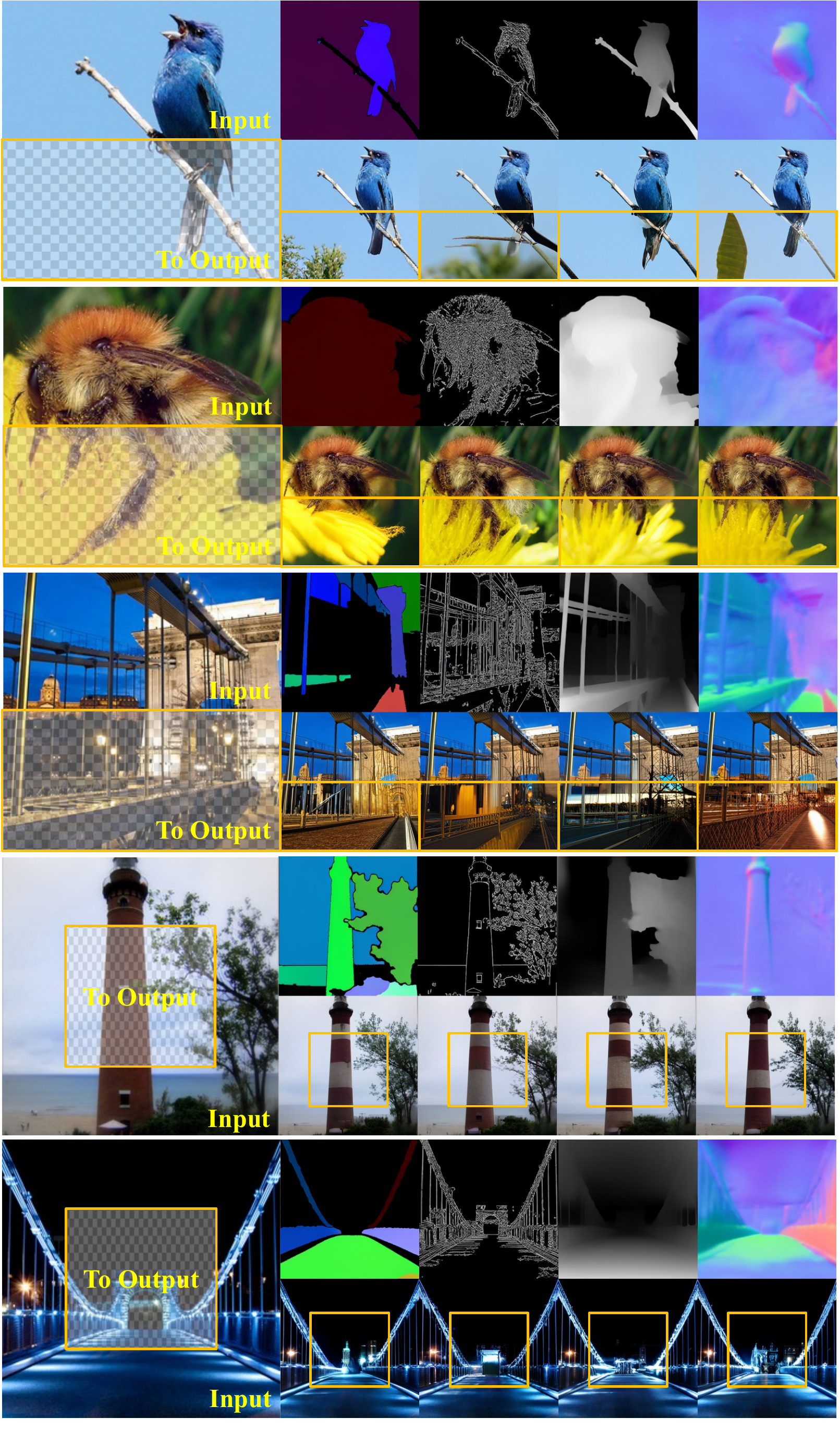}
    \caption{Qualitative visualization for image/control inpainting task. The {\color{yellow}{yellow}} boxes denote the predicted images/controls.}
    \label{fig:inpaint_supp}
\end{figure}

\begin{figure}[t]
    \centering
    \includegraphics[width=\linewidth]{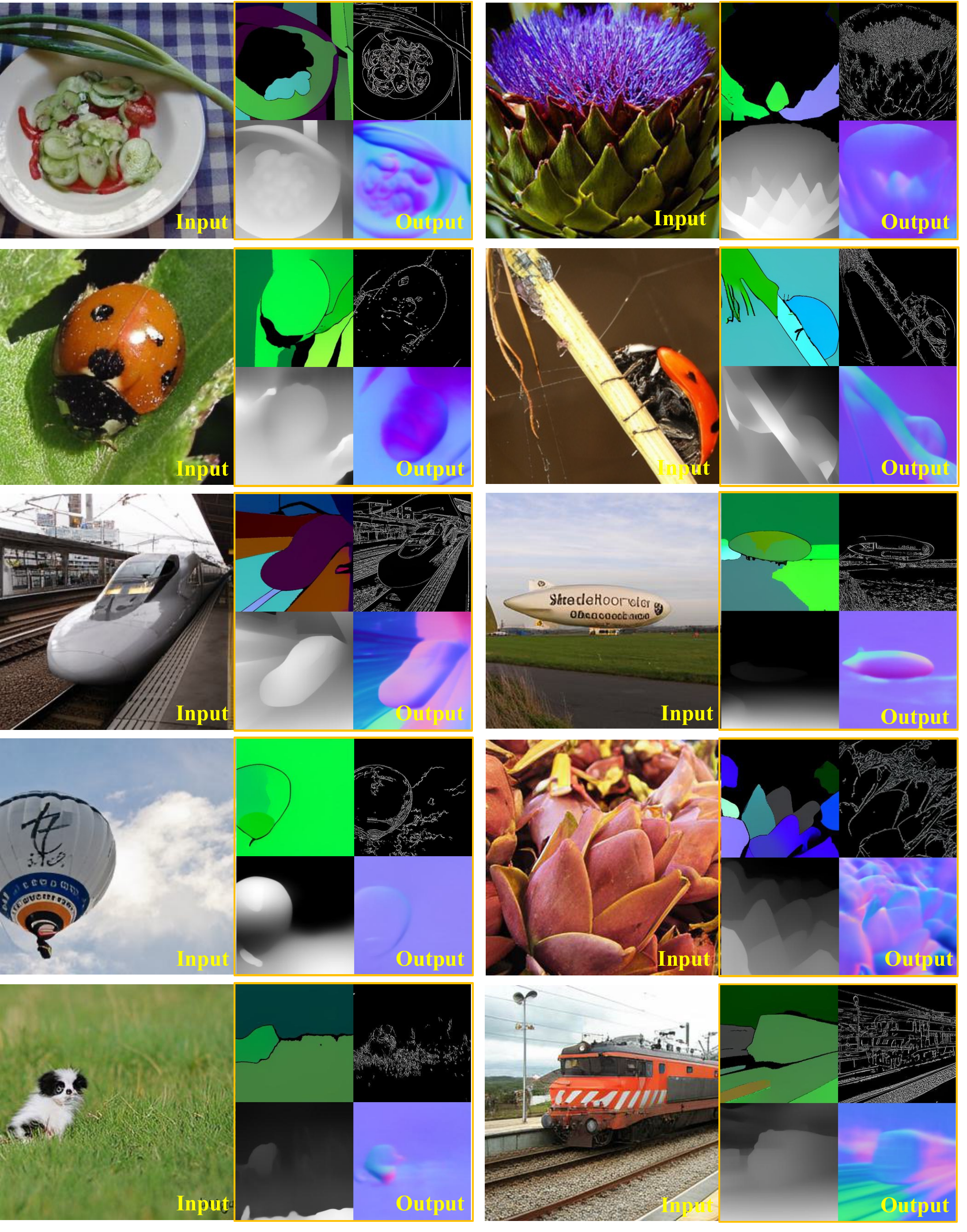}
    \caption{Qualitative visualization for image understanding task. The {\color{yellow}{yellow}} boxes denote the predicted controls.}
    \label{fig:percep_supp}
\end{figure}

We demonstrate more qualitative visualization for joint control-image generation (\cref{fig:joint_supp}), image/control completion (\cref{fig:inpaint_supp}), image perception (\cref{fig:percep_supp}), conditional image generation (\cref{fig:cgen_supp}) and unseen control-to-control generation (\cref{fig:zero_supp}).

\begin{figure}[t]
    \centering
    \includegraphics[width=\linewidth]{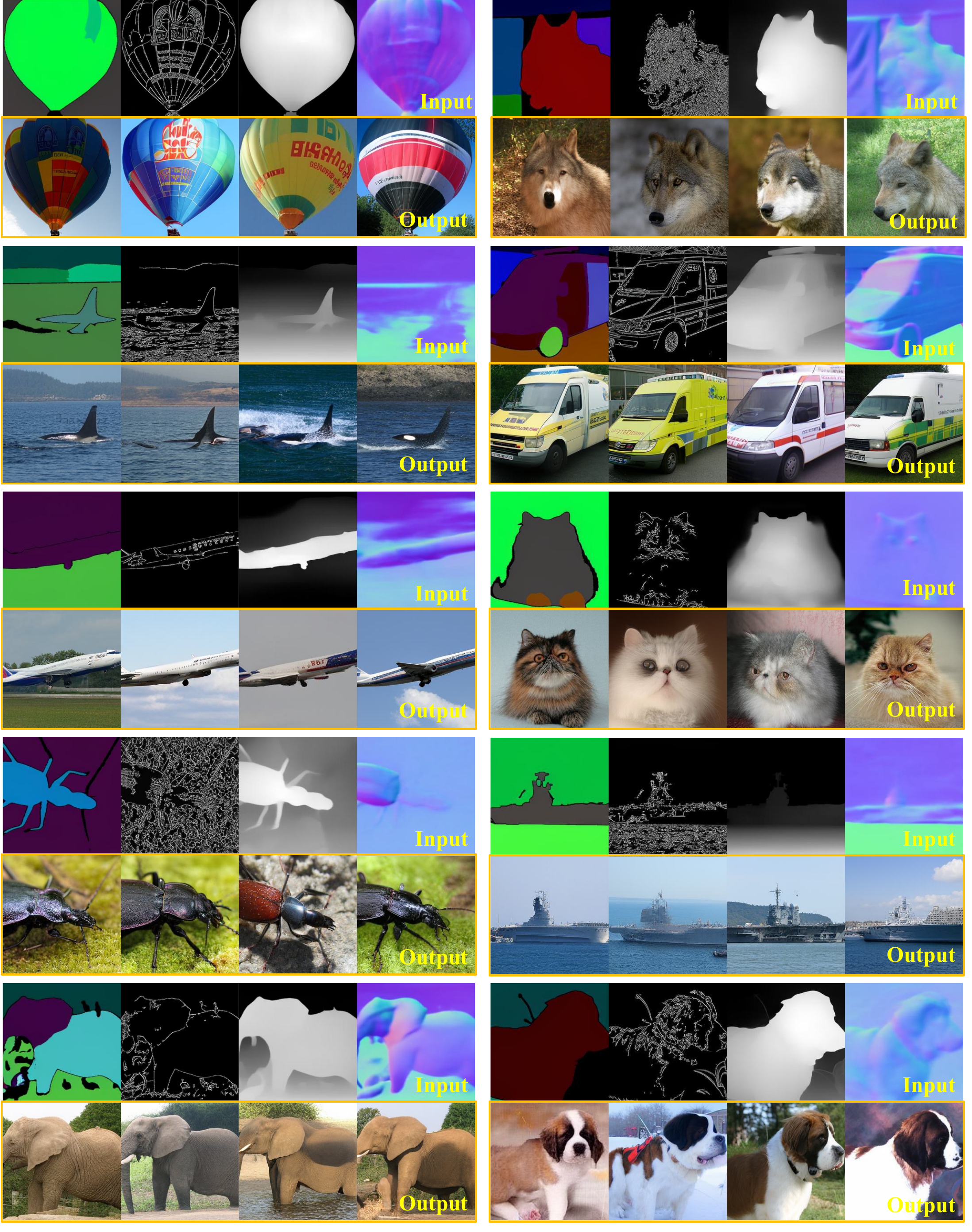}
    \caption{Qualitative visualization for conditional image generation task. The {\color{yellow}{yellow}} boxes denote the predicted images.}
    \label{fig:cgen_supp}
\end{figure}


\end{document}